\documentclass{article}

\usepackage{microtype}
\usepackage{graphicx}
\usepackage{subfigure}
\usepackage{booktabs} 
\usepackage{tcolorbox}
\usepackage{xcolor}
\usepackage{amsmath}
\usepackage{amssymb}
\usepackage{cases}
\usepackage{float}
\usepackage{booktabs}
\usepackage{makecell}
\usepackage{array}
\usepackage{courier}
\usepackage[scaled=0.94]{inconsolata}
\renewcommand{\texttt}[1]{{\fontfamily{zi4}\selectfont #1}}

\usepackage{hyperref}

\usepackage[accepted]{mlsys2025}
\usepackage{xcolor}
\definecolor{lightgray}{gray}{0.85}
\usepackage{fontawesome5}
\usepackage{graphicx}
\usepackage{subcaption}
\usepackage{xcolor}

\mlsystitlerunning{FarSkip-Collectives: Unhobbling Blocking Communication in Mixture of Experts Models}

\begin{document}

\twocolumn[
\mlsystitle{FarSkip-Collective: Unhobbling Blocking Communication in Mixture of Experts Models}



\mlsyssetsymbol{equal}{*}

\begin{mlsysauthorlist}
\mlsysauthor{Yonatan Dukler}{to}
\mlsysauthor{Guihong Li}{to}
\mlsysauthor{Deval Shah}{to}
\mlsysauthor{Jiang Liu}{to}
\mlsysauthor{Vikram Appia}{to}
\mlsysauthor{Emad Barsoum}{to}
\end{mlsysauthorlist}

\mlsysaffiliation{to}{Advanced Micro Devices Inc. (AMD)
}

\mlsyscorrespondingauthor{Yonatan Dukler}{yonatan.dukler@amd.com}

\mlsyskeywords{Machine Learning, MLSys}

\vskip 0.3in

\begin{abstract}

Blocking communication presents a major hurdle in running MoEs efficiently in distributed settings. To address this, we present \emph{FarSkip-Collective} which modifies the architecture of modern models to enable overlapping of their computation with communication. Our approach modifies the architecture to skip connections in the model and it is unclear a priori whether the modified model architecture can remain as capable, especially for large state-of-the-art models and while modifying all of the model layers. 
We answer this question in the affirmative and fully convert a series of state-of-the-art models varying from 16B to 109B parameters to enable overlapping of their communication while achieving accuracy that is comparable with their original open-source releases. For example, we convert Llama 4 Scout (109B) via self-distillation and achieve average accuracy within 1\% of its instruction tuned release averaged over a wide range of downstream evaluations. 
In addition to demonstrating retained accuracy of the large modified models, we realize the benefits of FarSkip-Collective through optimized implementations that explicitly overlap communication with computation, accelerating both training and inference in existing frameworks.
For inference, we demonstrate 32.6\% speedup in Time To First Token when serving a converted DeepSeek-V3 architecture with expert parallelism in SGLang and achieve 97.3\% communication-computation overlap during the prefill stage. 
During training, our approach enables 88.9\% communication overlap of the all-to-all communication collectives when pre-training DeepSeek-V3 MoE layers with expert parallelism.
\end{abstract}
]



\printAffiliationsAndNotice{}  

\section{Introduction}
Mixture of Experts (MoE) models have emerged as the de-facto model architecture for leading large language models (LLMs) in recent years \cite{shazeer2017outrageously,fedus2022switch,dai2024deepseekmoe}. MoEs activate a sparse subset of their total parameters, usually via a mixture of experts layer to replace the dense MLP sub-block. The sparsity and reduced computational cost of MoEs makes them amenable to even larger parameter scaling with new open-source MoE models routinely exceeding 500B total parameters \cite{deepseekai2025deepseekr1incentivizingreasoningcapability,team2025kimi,meta2025llama4}. Because of the fixed memory capacity of compute accelerators, MoEs require even more distribution for training and inference setups as compared to their dense LLM counterparts \cite{deepseek2024v3}. \par
However, distributed training and inference is not without a cost as activations and model weights need to be communicated between devices quickly; especially when the communication operations are \emph{blocking}, where the communication can only commence at a particular stage of processing and the next operation in the compute graph relies on it. Blocking communication patterns result in \emph{exposed} idle time when the accelerator is not running computations; this commonly appears in popular parallelism techniques such as Expert, Sequence, and Tensor Parallelism \cite{narayanan2021efficientMegatron-LM,shazeer2017outrageously} and is especially tricky to overcome during inference. The new age of large and sparsely activated MoE architectures along with improved hardware computation speeds exacerbates these issues as communication becomes a relatively larger portion of the end-to-end workload \cite{zhao2025insights}.
\par
In this work we present a method to modify models' architectures to use available activations for the next computation at the onset of the communication call, which may be outdated or partially materialized, in order to avoid the blocking communication and start the next computation during the communication operation. We name our approach FarSkip-Collective as we start the next computation immediately using available activations and run the communication collective in parallel, far-skipping the communicated result to the residual of the next layer and making that activation available for future layers. By running communication overlapped with computation, as long as the duration of the computation leading up to the next residual is longer than the communication we avoid idle compute time. 

Mathematically, FarSkip-Collective is ``dropping'' connections of the network as the input to the next computation is now a residual which does not include the latest communicated output block. Characteristically, this can damage the capabilities of the model architecture. We therefore focus on evaluating whether the modified architecture connectivity can perform comparably with regular MoE connectivity and study the capabilities of models exhaustively over a wide array of evaluations and model scales. 

By developing FarSkip-Collective Self-Distill (FCSD) we answer in the affirmative, demonstrating that state-of-the-art open-source MoE models of scales ranging from 16B to 109B can be fully converted into FarSkip-Collective models with minimal loss of capabilities of the model demonstrated at varying scales \cite{qwen3technicalreport,meta2025llama4,deepseekv2}.
 FCSD is a simple yet effective knowledge distillation recipe we identified via a systematic study which can be applied to any model with the absence of a powerful teacher. In addition, when ablating the FarSkip-Collective architecture by pretraining from scratch, we observe on-par performance at the 16B model scale, further corroborating the viability of the architecture.

Independently of this work, we became aware of recent works exploring similar approaches, modifying the model architecture and running computations with either ``outdated'' \cite{zhang2025ladder} or ``partial'' \cite{prabhakar2024kraken,lamprecht2025tensor} activations to overlap communication. 
In contrast with our work, existing approaches focus has been limited to tensor-parallelism in dense models which are not designed for the communication patterns of MoEs.
Such works have also only studied the problem of model capabilities at order of magnitude smaller-scale models or achieved only partial modification of the model layers. It therefore remained unclear whether FarSkip's modified connectivity can be applied to overlap communication in all layers of an MoE and continue to perform at the scale of today's state-of-the-art MoE LLMs. It is not uncommon for model architecture modification to show promise at a smaller size but scale poorly when studied at frontier-LLM scale with more challenging tasks. We find the results with FCSD encouraging in that even at the 100B+ model scale over a wide range of generation and likelihood-based tasks, FarSkip-Collective can achieve within 1\% on average from the original model's accuracy.
\par
Just because the new model architecture obviates dependencies in the model that regularly lead to blocking communication, it does not imply the architecture will automatically overlap computation and communication in practice if not implemented carefully. To this end, we realize the overlapping opportunities of the models by developing performant and overlapped implementations for training and inference which we have tested extensively. For training, we develop on top of Megatron-LM \cite{narayanan2021efficientMegatron-LM} and Primus \cite{AMD_AGI_Primus}. We achieve 88.4\% computation-communication overlapping of the Expert Parallelism communication (87.6\% in forward, 89.0\% in backward) using asynchronous execution of communication collectives and novel scheduling techniques at the PyTorch API layer. On the inference side, we implement our method on top of vLLM \& SGLang and integrate our approach with HIP/CUDA-graphs achieving up to 97.6\% communication overlap. 
Overall we implement our approach for wide use across different hardware and avoid low-level kernel optimizations in favor of more general implementation at the PyTorch layer. We open-source our overlapped implementations and provide easy integration with the upstream frameworks at \url{https://github.com/AMD-AGI/FarSkip-Collective}.

\begin{figure*}[ht]
    \centering
    \includegraphics[width=0.8\textwidth]{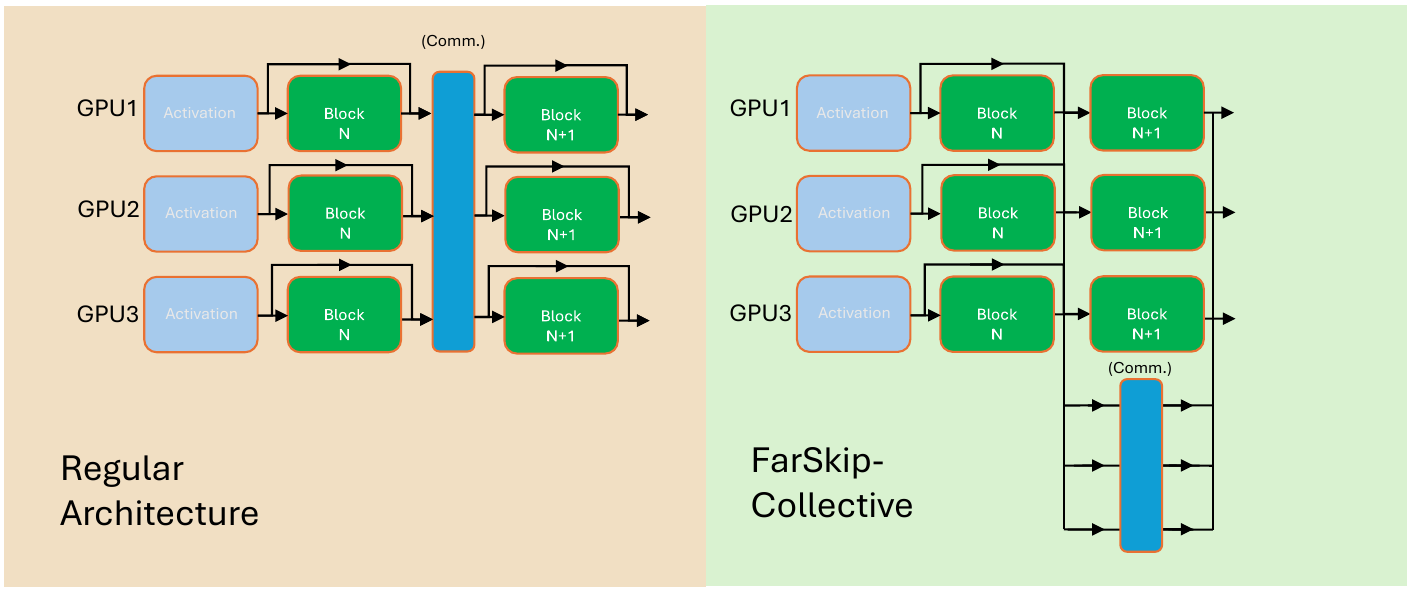}
    \caption{FarSkip-Collective modifies the connectivity between sub-blocks to avoid waiting for communication collectives. Computation continues with available activations, partial (e.g., Block N output) or outdated (e.g., Activation).}
    \label{fig:yourlabel}
\end{figure*}

\begin{figure*}[ht]

\end{figure*}

We summarize our contributions as follows:
\begin{itemize}
    \item We present FarSkip-Collectives, a method to convert the execution dependency of model layers that eliminates blocking communication patterns in MoEs, thereby allowing inference and training speedups.

    \item We demonstrate at the 100B+ parameter scale that models using the FarSkip-Collective architecture modifications retain the capabilities of modern transformer blocks while speeding up their execution in distributed settings. In particular we fully convert the Llama 4 Scout MoE (109B) while observing only small drop in model performance of 1\% on average compared with the instruction-tuned open-sourced model release. (In the case of pre-training from scratch we demonstrate an average drop in performance of 0.3\% when ablating the DeepSeek-V2-Lite architecture (16B)).
  
    \item We develop FCSD, an efficient and general knowledge self-distillation recipe to convert existing LLMs into FarSkip models and demonstrate it by converting DeepSeek-V2 Lite (16B), Qwen-3-30B-MoE (30B), and Llama 4 Scout (109B) with $<$ 10B training tokens.

    \item For large-scale training, we integrate our method into Megatron-LM and achieve 88.9\% communication overlap for the previously blocking all-to-all communication responsible for MoE expert parallelism.

    \item For model serving, we develop an optimized implementation of FarSkip in SGLang and vLLM that overlaps the communication for distributed inference. 
    For example, for the modified Llama-4 Scout model, we achieve 18.5\% speedup in Time To First Token.

     \end{itemize}
The rest of the paper is organized as follows, in Section \ref{sec:background} we present background followed by our approach in Section \ref{sec:framework} and explicit optimized implementation of the method in Section \ref{sec:overlapping}. In Section \ref{sec:experiments}, we present our experimental results and review related works in Section \ref{sec:related} followed by conclusion (Section \ref{sec:conclusion}).

\section{Background}\label{sec:background}
\subsection{MoE parallelism at training and inference}
Two of the key parallelism techniques for MoE training and inference are Tensor and Expert Parallelism.

\paragraph{Tensor Parallelism} In Tensor Parallelism (TP), an MLP or a Self-Attention sub-block will be split by slicing the sub-block's weight matrices evenly across their columns or rows. Let $A \in \mathbb{R}^{B\times d}$ be a model input activation for a modern MLP layer of the form
\begin{equation*}
    \text{MLP}(A) =\sigma(AW_1^{\top}\cdot g(AW_2^{\top}))W_3^{\top},
\end{equation*}
with $g,\sigma$ being entrywise non-linearities, $W_1,W_2 \in \mathbb{R}^{c \times d}, W_3 \in \mathbb{R}^{d\times c}$.
TP of size $k$ will split the matrices into
\begin{align*}
W_{\{1,2\}}^i &= \big(W_{\{1,2\}}\big)_{[i*c/k:(i+1)*c/k,:]} \in \mathbb{R}^{(c/k) \times d}, \\
W_3^i &= \big(W_3\big)_{[:,i*c/k:(i+1)*c/k]} \in \mathbb{R}^{d \times (c/k)},
\end{align*}
for $0\le i \le k-1$.
Then computation of each TP rank can run independently until the end of the sub-block where an all-reduce communication-collective is applied to construct the final output,
\begin{gather}
    \text{MLP}(A)_i = \sigma(A W_1^{i\top} \cdot g(A W_2^{i\top})) W_3^{i\top}, \\
    \text{MLP}(A) = \text{all-reduce}\big(\text{MLP}(A)_i\big).\label{eq:mlp_tp}
\end{gather}
The selection of Column-Parallel $W_{\{1,2\}}^i$ split followed by Row-Parallel $W_3^i$ split makes it possible to only apply communication once at the end of the layer.

For multi-head self-attention (or efficient variants) implemented with TP, the computation is decomposed into independent computations partitioned across the different attention heads; this allows for a similar implementation employing Column-Parallelism ($Q,K,V$) followed by Row-Parallelism ($O$) and a single all-reduce.

\paragraph{Expert Parallelism} The key parallelism component of MoE layers is Expert Parallelism (EP). An MoE layer with $E$ experts generalizes the MLP as
\begin{equation*}
    \text{MoE}(A) = \sum_{j=1}^{E} G(A)_j \cdot \text{MLP}^j(A),
\end{equation*}
where $G(A) = s(A W_R^\top)$ is a linear classification layer followed by a sparse router selection function $s()$ activating only a subset of the $\text{MLP}^j$s. Here $\text{MLP}^j(A)$ refers  to a distinct ``expert'' for $1\le j \le E$. With EP of rank $k$, subsets of $\sim E/k$ experts will be distributed across the $k$ parallel ranks. Unlike TP, during training different input activations will be mapped to the different experts based on the router selection $G(A)$. Mechanically, specific token vectors $A_{[l,:]} \in \mathbb{R}^{d}$ will be grouped and mapped to a subset of experts $P_l \subset \{1,\dots k\}$ requiring permutation of $A$ followed by an all-to-all collective that sends and receives data between the ranks according to the router-defined data-partition map.
\begin{equation*}
    A_i = R_i \times A \;\; \text{placed on rank } i \quad \text{(Dispatch)},
\end{equation*}
with $R_i\in \{0, 1\}^{B_i \times d}$ being the indicator of $G(A)$; this is referred to as ``Dispatch'' and will have different bandwidth requirements depending on factors such as the number of experts $E$ and the sparsity of $s$ (e.g., the assigned ``TopK'' value). After each expert receives its dedicated tokens and computes $\text{MLP}^j(A)$ of the relevant experts on its rank, a dual all-to-all collective, referred to as ``Combine'' is applied to aggregate and sum the routed-experts' activations back into the output activation of the MoE layer.
Modern MoE designs will also typically include ``shared-experts'' MLP layers that will run on all tokens in addition to the routed MoE experts.

Putting these together, typical training execution of an MoE transformer layer would follow 1) the attention sub-block (including layer-norm) 2) potential post-attention communication collective if TP or CP is enabled 3) compute the gating and router scores (duplicated on each rank) 4) initiate routed expert Dispatch communication 5) routed expert computation and potentially shared-experts computation, and finally 6) Combine collective operation to aggregate the routed-experts from the different ranks. This execution leads to three potential blocking communication bubbles: (a) post-attention blocking communication if TP/CP is used, (b) during Dispatch and (c) during Combine, although (c) may be partially overlapped if shared experts are present.

For inference, there are different approaches for MoE execution, with only some involving all-to-all  ``Dispatch + Combine'' \cite{deepseek2024v3}. In this work, we focus on the approach implemented by vLLM and SGLang where an all-reduce is used instead of the all-to-all collective and activations are replicated and indexed across the expert parallel ranks.

\subsection{Model Distillation}
FarSkip-Collective modifies the model architecture followed by self-distillation to recover the original model's capabilities. As a basic approach, one can simply fine-tune the model with high-quality data via Supervised Fine-tuning (SFT) according to
\begin{equation}
\mathcal{L}_{\text{SFT}}(\theta)
= -\,\mathbb{E}_{(x, y) \sim \mathcal{D}}
\left[
\sum_{t=1}^{|y|}
\log p_\theta\!\left(y_t \mid x, y_{<t}\right)
\right],\label{eq:sft}
\end{equation}
where $\theta$ denotes the model parameters, and $(x, y) \sim \mathcal{D}$ are input--output supervision pairs with $|y|$ output tokens in the pair.

When converting the model with a target model in mind (e.g., in our case aiming to recover the original model), one may train with knowledge distillation which is defined against a fixed teacher model $q$. Logit-based knowledge distillation optimizes $p_\theta$ to match the teacher's predictive distribution according to
\begin{equation}
\mathcal{L}_{\footnotesize{\text{KD}}}(\theta)
\!=\!\mathop{\mathbb{E}}_{x\sim\mathcal{D}}
\Bigg[
\sum_{t=1}^{T}
\mathrm{KL}\!\big(q(\cdot\mid x,y_{<t})\,\|\,p_\theta(\cdot\mid x,y_{<t})\big)
\Bigg].\label{eq:kl}
\end{equation}
In addition to an objective on the model outputs, one may also aim to increase alignment of a model to a teacher model $q$ by aligning the model with the intermediate representations of the teacher \cite{yang2025zebra} as
\begin{equation}
\mathcal{L}_{L2}(\theta) = \sum_{i=1}^{L} \|\, o_i(\theta)- t_i \,\|_2^2, \label{eq:l2}
\end{equation}
where $o_i$ and $t_i$ denote the matching hidden activations of the student and teacher models, respectively, over $L$ layers.

\begin{figure*}[ht]
    \centering
    \begin{minipage}[t]{0.65\linewidth}
    \vspace{0.0pt}  
    \centering
    \includegraphics[width=1.0\textwidth]{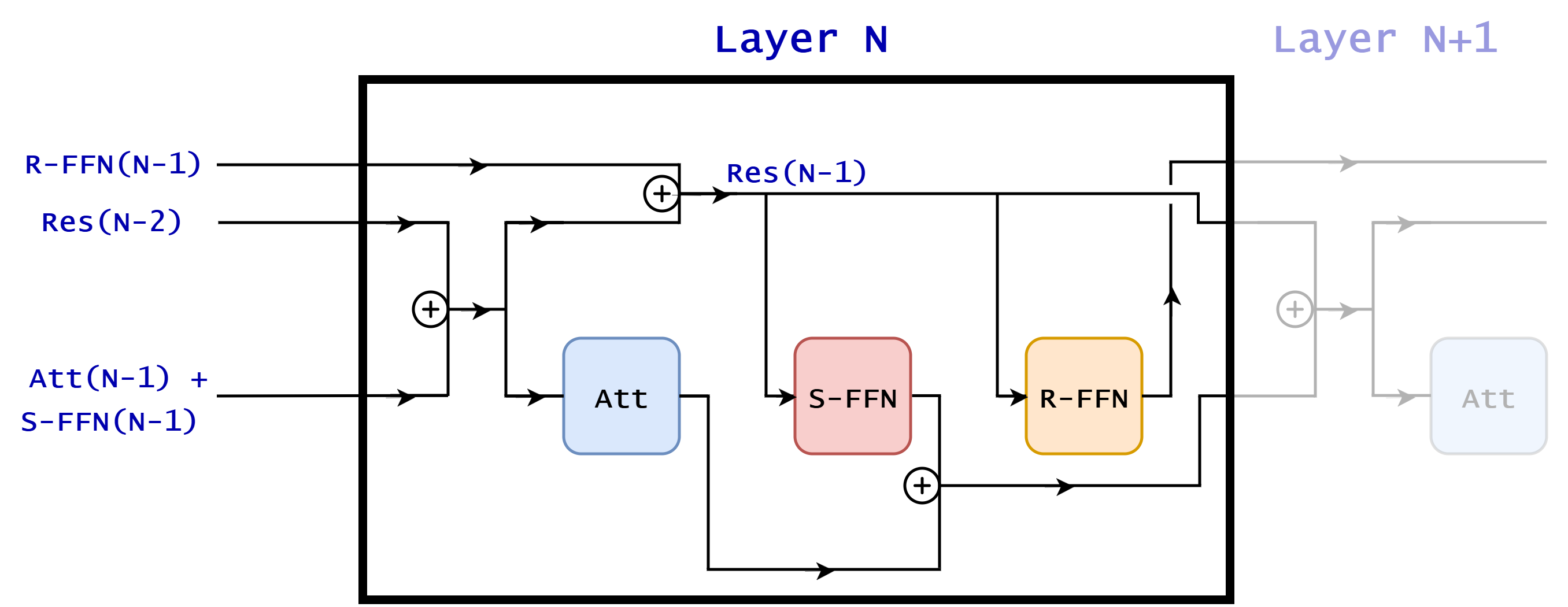}
    \caption{FarSkip-Collective MoE block diagram. The attention (Att) block's input corresponds to partial output from the previous layer. Both the shared and routed experts (S-FFN, R-FFN) parts of the MLP block receive outdated input (residual of layer N-1). Each input is at most a single layer behind while communication can be fully overlapped. Since the output of attention and shared-experts are used together, their collectives may be combined to reduce bandwidth.}
    \label{fig:farskip_diag}
    \end{minipage}%
    \hfill
    \begin{minipage}[t]{0.32\linewidth}
    \vspace{0pt}   
        \centering
    \includegraphics[width=1.052\columnwidth]{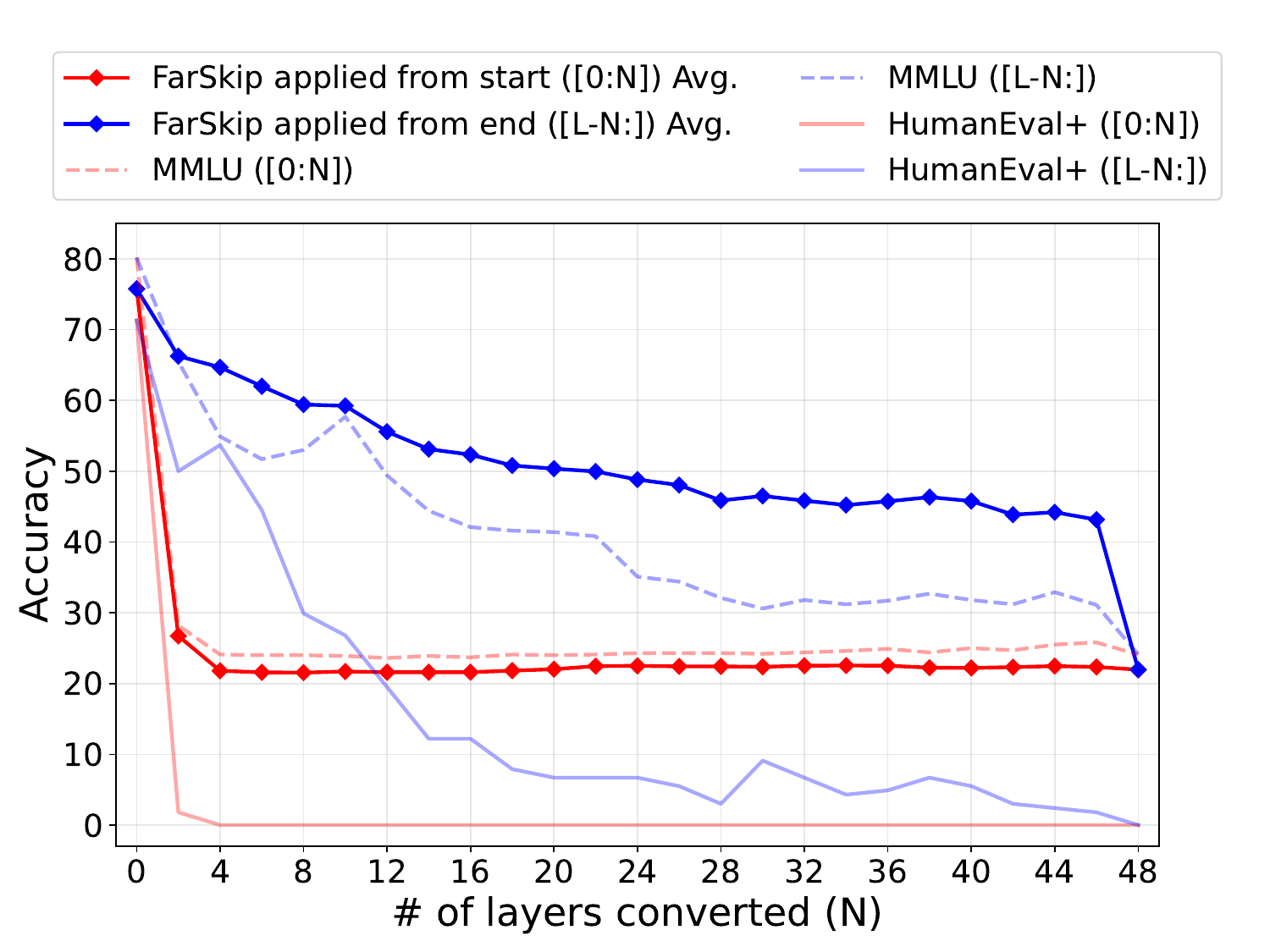}
    \caption{Accuracy of Qwen-3-30B-MoE modified with $N$ FarSkip-Collective layers \underline{without training}. Replacing the last $N$ layers (blue) and first $N$ layers (red).}\label{fig:sensitivity}
    \end{minipage}
\end{figure*}

\section{FarSkip-Collective Framework}\label{sec:framework}

Modern networks use residual connections, meaning that self-attention, MLP, and MoE sub-block outputs are incrementally added to a residual activation stream. Denote the output activation of a network after $k$ layers as $o_k$, and the $i$th sub-block (layer) of the network as $f_i$. The output $o_k$ is computed as
\begin{equation}
    o_k = f_1(o_0) + f_2(o_1) + \dots + f_k(o_{k-1}).\label{eq:reg}
\end{equation}
For large models, producing $f_k$ can involve blocking communication which stops $o_k$ from being used as an input to $f_{k+1}$ until $o_k$ is communicated -- leading to idle computation resources. We propose to use an available activation instead, denoted as $o_k^*$, to be used as input to $f_{k+1}$ and compute the next layer while the communication collective is producing $o_{k}$. $o_{k+1}$ will be updated with $o_{k}$ once it is ready; however, now the communication of $o_k$  can be ``far-skipped'' and overlapped over the duration of the computation of $f_{k+1}(o_k^{*})$ by using $o^*_k$ as the input to the block.
\begin{equation}
\begin{aligned}
o_{k+1} &= o_{k} + f_{k+1}(o^*_{k}) \\
    &= f_1(o_0) + f_2(o^*_1) + \dots + f_{k+1}(o^*_{k})
\end{aligned}
\label{eq:farskip}
\end{equation}
We consider two options for $o_k^*$,
\begin{numcases}{}
    o_k^* = o_{k-1} \quad \text{(outdated)} \tag{8a}\label{eq:outdated} \\
    \text{or} \notag \\
    o_k^* = o_{k-1} + f^*_k(o_{k-1}^*) \quad \text{(partial)} \tag{8b}\label{eq:partial}
\end{numcases}
\addtocounter{equation}{1}

$f^*_{k}$ denotes part of the activation of block $f_k$ which is ready before the collective, e.g., $\text{MLP}_i(A)$ in Eq. \ref{eq:mlp_tp}.
In both cases, the output activation $o_{k}$ will consist of the same number of blocks as before, but the difference will be in terms of the input activation into each $f_i$. Crucially the input to $f_{k+1}$, $o_k^{*}$, has access to all of the previous block outputs except for the full representation of block $f_{k}$; that is all future layers $f_{j}$ for $j\ge k+2$ will have access to the full $f_{k}$ outputs. 

When developing FarSkip-Collective for MoEs we balance two competing goals: minimize the number of activations dropped from each $o_k^*$ while maximizing possible overlap. Specifically for the MoE settings, we consider the possible communication patterns of the attention block, and shared and routed experts parts of the MLP block. The routed-experts are parallelized with EP which for training will require Dispatch and Combine, whereas the shared-experts and attention computations are not router conditional and tend to have compatible parallelisms (e.g. replication / TP / CP). When developing the FarSkip-connectivity we aim to combine the latter two's communications to reduce bandwidth, while at the same time overlapping the routed-experts communication. 

To this end we optimize the execution order and connectivity by considering 1) the routed-experts communication and 2) attention and shared-experts communication (if not replicated). 
The most minimal activation dropping scheme that allows overlapping both involves using the partial formulation for attention since we can prepare the non-routed activation of the MLP without blocking the attention computation. Then for the MLP, use the outdated formulation which can not depend on the previous attention block while remaining overlapped. We present this connectivity diagram in Fig.~\ref{fig:farskip_diag} demonstrating the input and output to each computation.

Mathematically let $o_k^{*}(\text{attn})$ be the input to the $k$th attention sub-block, then the partial activation
\begin{equation*}
    \text{attn-in}_k\!:=\!o_k^{*}(\text{attn})\!=\!o_{k-2} + \text{attn-out}_{k-1} + \text{shared-exp-out}_{k-1}
\end{equation*}
serves as the input.
For the MLP block input (shared and routed experts), $o_k^{*}(\text{mlp})$ is the outdated activation
\begin{equation*}
        \text{mlp-in}_k :=o_k^{*}(\text{mlp}) = o_{k-1}.
\end{equation*}
Compared with $\text{attn-in}_k$, $\text{mlp-in}_k$ can be decomposed as $o_{k-1} = \text{attn-in}_k + \text{routed-exp-out}_{k-1}$ with the routed-expert activations added after Combine finishes.

The FarSkip-Collective connectivity has the property that attention computation is not dependent on the preceding routed-experts which means the Combine operation of the previous routed-experts can be overlapped with the first part of the attention computation (and shared-experts computation depending on the parallelism). In addition the MLP input is independent of the attention output which means that the Dispatch can follow the Combine immediately after computing the new router scores and then run overlapped with the second part of the attention computation. Hence FarSkip-Collectives combines the outdated and partial formulations for $o_k^{*}$ to achieve overlap with provable minimal activation dropping for the MoE layer. We provide the details of the execution and overlapping of FarSkip-Collective in the next section. 

We also leave more aggressive multi-block variants of ``far-skipping'' as future work, which may be useful if the communication significantly exceeds the duration of the full sub-block's computation time, for example in the case of extremely sparse and large-scale MoEs or different hardware paradigms.

\subsection{Distilling existing models with the FarSkip-Collective Framework}
The FarSkip-Collective method modifies the architecture connectivity without changing the model's parameter layout or dense compute operations, making it possible to execute an existing checkpoint with FarSkip-Collective connectivity using the same kernels with relatively few modifications to the model definition. In Fig.~\ref{fig:sensitivity}, we give a simple demonstration of this by loading the original Qwen-3-30B MoE model checkpoint into models with various numbers of FarSkip-Collective layers activated and evaluate its performance on different benchmarks. Without re-training, we observe that as we increase the number of converted layers, the model performance degrades considerably, and the model achieves random baseline accuracy on MMLU and 0\% on HumanEval+ when fully converted. This is unsurprising, as we pass different input activations than the ones the model was trained with, leading to out-of-distribution outputs. 

We, however, show that by continuing training the original checkpoint via Knowledge Distillation \eqref{eq:kl} using typical instruction tuning data, we are able to recover the original model's performance in a small fraction of the compute needed to retrain it from scratch with the FarSkip-Collective architecture (${\sim}100 -1000{\times}$ cheaper). We systematically study different approaches for the distillation training which we present in Tab.~\ref{tab:training_comparison} and find that using KL-based knowledge distillation with the original model as the teacher (self-distillation) performs best or on par as compared to the different approaches we tested. We also study the effect of different aspects such as the batch-size and learning rate and find that they also contribute to the final model performance and training stability, which culminates in our simple and robust ``FarSkip-Collective Self-Distillation'' (FCSD) recipe to convert any MoE model into the FarSkip-Collective connectivity. Self-distillation to the original model's logits provides granular signal that better aligns with the existing representations. Training using the logits as the training signal, also reduces the dependency on meticulous and high quality instruction tuning data as the FarSkip-Collective model is instead trained to align with the original model's predictions rather than potentially low quality data itself. On the flip side this approach will also essentially cap the final performance of the student model to that of the teacher model especially since they are of the same size. We however find FCSD attractive in its generality since unlike other distillation approaches we do not rely on strong external teacher models and training with FCSD gives strong performance under 10B tokens.

Since the modified architecture only modifies the connectivity of the original model, (i.e., all the model parameters have the same shapes etc.) FCSD is effective in reducing the distance from the original model, nonetheless later in training KL may lead to training instabilities as small discrepancies between the teacher and student model lead to occasional large gradients and training instabilities. We tested different approaches to overcome this but find that training FCSD with early stopping enables us to avoid the issue. For the early stopping validation we use the MBPP+\cite{evalplus} dataset as a fast proxy for detecting instabilities and evaluate every 1000 training steps with a patience of 20 evaluations and performance delta of $2\%$. MBPP+ provides for quick evaluation and being a code-generation dataset it is sensitive to damaging distribution shifts caused by training instabilities. When evaluating conversion using a direct SFT objective as a baseline we apply the same early stopping procedure for fair comparison.

\begin{figure*}[ht]
    \centering
    \includegraphics[width=0.75\textwidth]{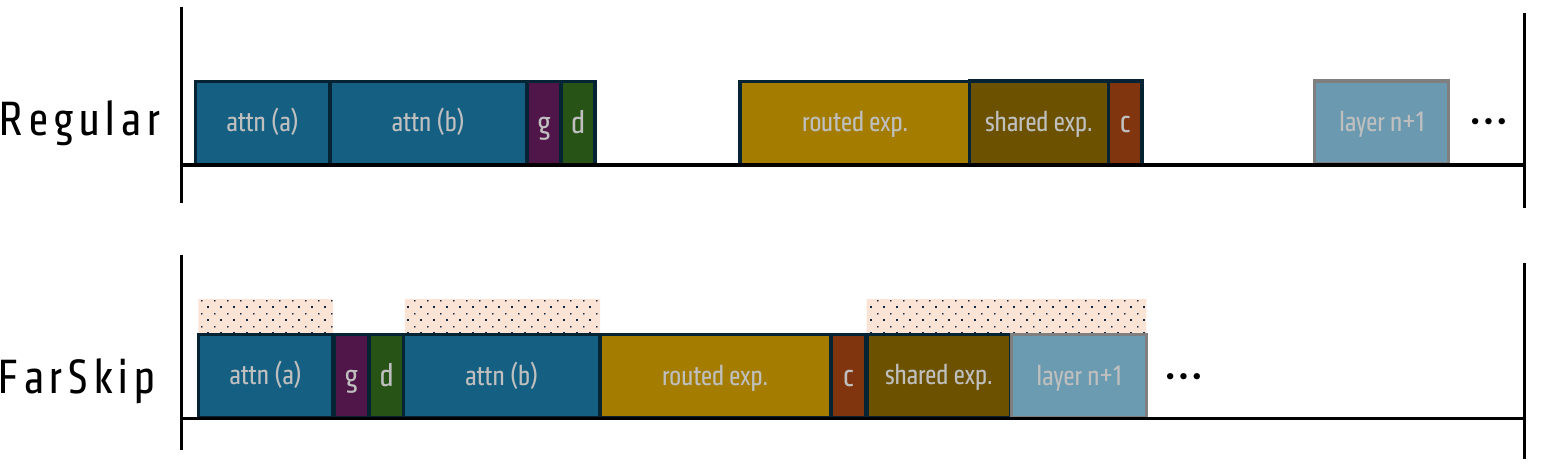}
    \caption{FarSkip-Collective MoE layer main operator execution. \emph{g,d} and \emph{c} refer to gating+routing, Dispatch start, and Combine start respectively. For communication operations, we only denote the starting point of the operation. The overlapping window enabled by FarSkip-Collective is illustrated with the shaded area above the operators. }
    \label{fig:moe_execution}
\end{figure*}

\section{Explicit overlapping of FarSkip models}\label{sec:overlapping}
Modern GPUs, equipped with hundreds of independent Compute Units (Streaming Multiprocessors), can process multiple Queues (Streams) of kernels independently by scheduling work on different sets of processing units at the same time \cite{zhao2023pytorch,pytorch_async_tp_2024,deepseek2024v3}. Both computation and communication operations utilize compute units to run operations, with communication operations only utilizing a fraction of the total available units, allowing for overlap with computation. Computation-communication overlap, however, requires dedicated implementation, and aside from standard patterns, modern frameworks such as PyTorch and JAX will not accomplish this automatically.
Therefore, even though the modified FarSkip-Collective model will logically facilitate parallel and non-blocking flow through the computational graph, without explicit implementation, the models will not automatically overlap communication with computation. 
Below, we describe our optimized and extensively tested explicitly-overlapping implementation of the FarSkip-Collective framework. 
As a design choice, we aim to make our implementation generalizable and as hardware-independent as possible by sticking to the framework level as opposed to lower-level kernels or Triton. In particular to enable scheduling of non-blocking communication calls, we rely on \texttt{torch.dist}'s \texttt{async\_op=True} parameter or using the \texttt{torch.cuda.Stream()} context that enables more granular control of scheduling of kernels on the non-main queue. Note that by overlapping operations, one diverts some of the processing units, which can lead to unavoidable slowdown in computations as compared to when they are solely executed on the hardware.

\subsection{Training}
For training, we consider MegatronLM GPT training with MoE layers. The basic settings of training include shared-experts, Multi-head Latent Attention (MLA) and running with EP while replicating attention (no TP), following the DeepSeek's V3 model training recipe. Note however that FarSkip also allows for full overlapping of mixed EP parallelisms during training. Regularly in this setup, the all-to-all collective will lead to two communication bubbles as part of Dispatch and Combine, appearing both during the forward and backward pass of each layer. With the blocking dependency removed by our modified architecture, we modify the execution order discussed in Section~\ref{sec:background}. In particular, we split the attention sub-block calculation into two parts: a) MLA preparation of ($q,k,v$) and b) core-attention + output projection. This enables us to easily launch a communication kernel asynchronously between the two parts and then immediately continue the attention calculation. For this DeepSeek model setup, we execute the FarSkip-Collective MoE layer forward pass as 1) attention part (a) computation, 2) synchronize Combine if last layer was a FarSkip-Collective MoE layer, 3) compute MoE gating and router scores, 4) initiate Dispatch (async mode will return immediately), 5) compute attention part (b). At this point, attention computation finished and the routed tokens should be dispatched; we 6) synchronize the Dispatch collective followed by computing the routed-experts and then 7) initiate Combine (async mode), lastly followed by 8) running shared-experts computation. We provide a visual demonstration of this layer execution as compared with the regular operation flow in Fig. \ref{fig:moe_execution}. With FarSkip we maximize the window of all-to-all communication overlap during the forward call as
\begin{equation}
\begin{aligned}
    T_{\text{Dispatch}} + T_{\text{Combine}} &\le T_{\text{overlapped computation}} \\
    &= T_{\text{layer}} - (T_{\text{Routed-Experts}} + T_{\text{Gate}}).
\end{aligned}\label{eq:overlap_potential}
\end{equation}
The only computations of the layers that cannot be overlapped with Dispatch and Combine communication are the routed-experts and gating operator. This is because with the modified architecture the routed-experts consume input from the previous Dispatch and produce outputs that will serve as the input for the next Combine call. Similarly, the gating operator will consume input from the previous Combine call and produce outputs that will serve as an input to the next Dispatch call. 

\par For the backward pass, we would like to overlap the Combine and Dispatch gradient calls which will also trigger blocking all-to-all communication collectives. If run naively, one will need to run the backward communication outside of async mode as its outputs will need to be synchronized for the next gradient in the graph, making the communication blocking again. The standard approach to avoid this is to explicitly control the operator ordering in backward by using a custom \texttt{torch.autograd.Function} for the \emph{entire} MoE transformer block layer computation. Implementing just Dispatch and Combine with a custom backward or sub-parts does not enable one to define the synchronization points outside of it, which is needed for overlap. Implementing such a large layer's backward computation graph manually, however, is tedious and error-prone as each of the operations and weights needs to be wired correctly to their next input.\par 
Instead, we present two innovative techniques that together cleanly achieve overlap while continuing to rely on the automatic autograd for backward propagation. First, we achieve async-safe all-to-all backward communication functions that are both async and yet have synchronization points right before accessing their gradient results.  Second, we ``hijack'' the priority ordering of \texttt{torch.autograd} via the Sequence Number PyTorch autograd's internal implementation\footnotemark \footnotetext{See ``forward-backward correlation'' discussion of Sequence Number autograd internals in paragraph below the anchor {\scriptsize \href{https://docs.pytorch.org/docs/stable/autograd.html\#torch.autograd.profiler.emit_nvtx}{https://docs.pytorch.org/docs/stable/autograd.html\#torch.autograd.profiler.emit\_nvtx}}} to control the gradient node execution ordering and ensure the async-safe backward calls are seperated far apart to have sufficient overlap before their synchronization points are triggered. We provide an expanded explanation and details of our novel overlapped backward technique in Appendix \ref{appendix:backward_pass}.
Using our optimized implementation, we achieve an overall overlap of 88.4\% of the all-to-all communication time when training a DeepSeek-V2 Lite with EP8 parallelism as observed in Tab.~\ref{tab:overlap_megatron}. Note that the first all-to-all in backward and last all-to-all in forward cannot be overlapped as there are no additional computation candidates.

\subsection{Inference}
For inference, we implemented FarSkip in vLLM and later extended it to SGLang. SGLang and vLLM serve as modern LLM inference engines with TP, EP, and PP support for MoEs such as DeepSeek. Unlike other MoE EP implementations that use a pair of all-to-all collectives for Dispatch and Combine, in vLLM and SGLang model activations are replicated across the ranks but model weights including expert weights are still distributed via EP and TP. This approach eliminates the need for Dispatch and Combine and is implemented with all-reduce operations applied to the activations after the MLP layers finish. For the attention sub-block, vLLM and SGLang adopt a regular TP approach with an all-reduce collective and both of the all-reduce calls are regularly blocking as the activations are needed in the next layer.
\par
To implement FarSkip-Collective for the routed-experts MoE layer, we run the \texttt{all-reduce} in async-op mode and synchronize it only before the next MoE computations, as those activations are no longer needed for Dispatch and can be overlapped with attention computation. To reduce communication bandwidth, we present an optimized delayed approach for the communication of the attention outputs where usually communication appears in \texttt{RowParallelLinear} of the output projection. Instead we defer their communication and combine the attention and shared-experts activations into a single all-reduce call of the same message size by reducing the summed activation. This all-reduce call can then be overlapped with the routed-expert's computation. In inference with this parallelism settings, the overlappable window corresponds to the complement of the shared-experts computation ($T_{\text{Layer}} - T_{\text{shared Experts}}$). 
\par
For specialized attention such as MLA in DeepSeek models, prefill and generation will run different fused kernels, and we treat each case separately but defer and combine the all-reduce async-op call in each scenario.
To integrate FarSkip with HIP/CUDA-graphs we use graph-compatible communication API calls and use direct Python binding of NCCL (PyNCCL).
We test our inference pipeline using the self-distilled models fine-tuned with FCSD and observe that our distillation recovers the model performance in chat-based generation.
\begin{table*}[ht]
\caption{Original and distilled FarSkip-Collective model performance on downstream evaluation tasks.}
\label{tab:model_comparison}
\vskip 0.15in
\begin{center}
\begin{small}
\begin{sc}
\setlength{\tabcolsep}{3.4pt} 
\begin{tabular}{lcccccccccccccc}
\toprule
\scriptsize Model & \scriptsize Params & \scriptsize PIQA & \scriptsize ARC-E & \scriptsize ARC-C & \scriptsize HS & \scriptsize CSQa & \scriptsize WG & \scriptsize HEval+ & \scriptsize MMLU & \scriptsize OpenBook & \scriptsize GSM-8K & \scriptsize MBPP+ & \scriptsize Avg \\
\midrule
\scriptsize DeepSeek-V2-Lite (Original) & 16B & 80.1 & 80.2 & 53.8 & 80.8 & 69.1 & 72.1 & 40.2 & 56.8 & 45.2 & 70.1 & 60.8 & 64.5 \\
\scriptsize DeepSeek-V2-Lite (FCSD) & 16B & 79.9 & 78.9 & 50.0 & 76.9 & 70.1 & 68.4 & 41.5 & 50.5 & 41.8 & 64.2 & 59.8 & 62.0 \\
\scriptsize DeepSeek-V2-Lite (SFT) & 16B & 78.2 & 74.3 & 43.8 & 74.1 & 65.5 & 69.0 & 11.0 & 48.0 & 41.2 & 54.3 & 45.8 & 55.0 \\
\midrule
\scriptsize Qwen-3-30B MoE (Original) & 30B & 80.5 & 84.8 & 61.9 & 79.7 & 84.8 & 72.9 & 73.8 & 80.2 & 45.0 & 86.9 & 84.4 & 75.9 \\
\scriptsize Qwen-3-30B MoE (FCSD) & 30B & 80.4 & 83.3 & 58.5 & 77.2 & 84.9 & 74.0 & 73.2 & 74.0 & 42.8 & 87.6 & 74.4 & 73.7 \\
\scriptsize Qwen-3-30B MoE (SFT) & 30B & 77.8 & 69.4 & 44.9 & 75.6 & 68.9 & 65.6 & 0.6 & 63.1 & 41.4 & 76.0 & 71.7 & 59.5 \\
\midrule
\scriptsize Llama-4-Scout (Original) & 109B & 81.1 & 87.3 & 64.6 & 82.9 & 84.4 & 76.6 & 62.2 & 80.0 & 45.2 & 88.6 & 83.6 & 76.0 \\
\scriptsize Llama-4-Scout (FCSD) & 109B & 80.8 & 87.0 & 62.4 & 82.0 & 82.4 & 75.8 & 63.4 & 75.9 & 44.4 & 89.8 & 81.7 & 75.1 \\
\scriptsize Llama-4-Scout (SFT) & 109B & 80.7 & 80.3 & 52.4 & 80.0 & 72.0 & 76.2 & 14.0 & 69.7 & 43.8 & 78.6 & 73.5 & 65.6 \\
\bottomrule
\end{tabular}
\end{sc}
\end{small}
\end{center}
\vskip -0.1in
\end{table*}

\section{Experiments}\label{sec:experiments}

In this section we describe our experiments evaluating the model capabilities of FarSkip-Collective models, followed by evaluation of the FarSkip-enabled optimized overlapped implementation.

\subsection{Model Capabilities}

We present the main results of our distillation experiments in Tab.~\ref{tab:model_comparison}, where we consider three open-source state-of-the-art MoEs at different scales: DeepSeek-V2-Lite (16B-A3B), Qwen-3-30B MoE (30B-A3B), and Llama-4 Scout (109B-A17B). Each model's checkpoint corresponds to the instruction-tuned / chat version of the open-source model release. We apply FarSkip-Collective to all of the model's layers and train each model for up to 10B tokens of instruction tuning data. We train with standard settings using AdamW, cosine-annealing learning rate scheduler, and 1000-step warm-up period.
We use relativity large batch-size and learning rate with FCSD and run short sweeps to identify the best batch-size and learning rate for each model. In particular we conduct two sweeps for 2000 training steps each, first for batch-size selection among \{ $2^{16}, 2^{17},2^{18}$\} with a learning rate of 2e-5 followed by a learning rate sweep among \{2e-5,4e-5,8e-5\} where we use the training loss for selection. We observe rapid initial improvement on all benchmarks using the KL objective and further gradual improvement as training continues. We also observe occasional training instabilities where the distilled FarSkip-Collective model exhibits mode-collapse later in training. We tested different approaches to overcome this in Tab.~\ref{tab:training_comparison}, and resort to using early stopping with MBPP+. As a baseline conversion method, we test standard SFT training with the same training schedule and sweep selection for the batch-size and learning rate which is similar to the approach in \cite{zhang2025ladder}. In addition we apply the same early stopping as FCSD. Overall, SFT significantly underperforms the FCSD recipe and the resulting model exhibits catastrophic forgetting, particularly in generation tasks. 
With our knowledge distillation training, even for code generation task such as HumanEval+ which are more easily affected by distribution shifts, FarSkip-Collective models are able to achieve performance close to the original instruction-tuned checkpoint, demonstrating the inherent capacity of the modified architecture. Nonetheless we note that for some datasets there remains a gap from the original instructed tuned checkpoint, for example for Llama-4-Scout FSCD still results in 4.1 gap in MMLU performance vs. 10.3 gap with SFT. Overall, we believe additional scaling of FCSD, improved data mixtures and other techniques such as model merging can aid in closing any remaining performance gaps in the model conversion which we leave for future work.

In this vein, we report pre-training from scratch results in Tab.~\ref{tab:farskip_pretrain} and observe on par performance with 0.3\% gap on average, when conducting an architecture ablation with FarSkip-Collective while fixing all non-architectural aspects. In the pre-training settings we use a DeepSeek-V2-Lite model architecture (16B) and train with a 200B token budget, evaluating the final checkpoint. We further discuss the pre-training results in Appendix \ref{appendix:pre_training_results}.

\begin{table}[h]
\caption{Downstream performance of different training settings of FarSkip-Collective distillation. We evaluate different training settings and conversion settings training for 300M tokens.}
\label{tab:training_comparison}
\vskip 0.15in
\begin{center}
\begin{small}
\begin{sc}
\setlength{\tabcolsep}{3pt}
\begin{tabular}{lccccc}
\toprule
\scriptsize Model & \scriptsize ARC-C & \scriptsize HEval+ & \scriptsize MBPP+ & \scriptsize MMLU & \scriptsize Avg-11 \\
\midrule
\scriptsize Original & \footnotesize 53.8 & \footnotesize 40.2 & \footnotesize 60.8 & \footnotesize 56.8 & \footnotesize 64.5 \\
\midrule
\scriptsize KL (Far 50\%) & \footnotesize 51.0 & \footnotesize 42.7 & \footnotesize 60.6 & \footnotesize 57.3 & \footnotesize 63.9 \\
\scriptsize KL (Far 75\%) & \footnotesize 46.6 & \footnotesize 22.0 & \footnotesize 47.4 & \footnotesize 48.3 & \footnotesize 56.3 \\
\scriptsize KL (Far 90\%) & \footnotesize 46.0 & \footnotesize 20.1 & \footnotesize 38.4 & \footnotesize 43.0 & \footnotesize 52.5 \\
\scriptsize KL (Far 100\%) & \footnotesize 42.0 & \footnotesize 16.5 & \footnotesize 29.9 & \footnotesize 38.7 & \footnotesize 48.6 \\
\midrule
\scriptsize KL & \footnotesize 42.0 & \footnotesize 16.5 & \footnotesize 29.9 & \footnotesize 38.7 & \footnotesize 48.6 \\
\scriptsize KL + inter. L2 & \footnotesize 41.6 & \footnotesize 14.0 & \footnotesize 26.2 & \footnotesize 35.8 & \footnotesize 46.0 \\
\scriptsize SFT & \footnotesize 37.4 & \footnotesize 8.5 & \footnotesize 28.3 & \footnotesize 34.0 & \footnotesize 44.7 \\
\scriptsize KL \tiny{\faSnowflake}\scriptsize{ embed.} & \footnotesize 42.3 & \footnotesize 17.7 & \footnotesize 30.7 & \footnotesize 38.4 & \footnotesize 48.6 \\
\scriptsize KL $~4\times$bs (1.2B tokens) & \footnotesize 42.6 & \footnotesize 18.9 & \footnotesize 33.1 & \footnotesize 39.9 & \footnotesize 50.3 \\
\bottomrule
\end{tabular}
\end{sc}
\end{small}
\end{center}
\vskip -0.1in
\end{table}

\begin{table}[ht]
\caption{Pretraining model performance of Regular and FarSkip-Collective MoE using DS-V2-Lite architecture (16B) trained on 200B tokens.}
\label{tab:farskip_pretrain}
\vskip 0.15in
\begin{center}
\begin{small}
\begin{sc}
\setlength{\tabcolsep}{5pt}
\begin{tabular}{lcc}
\toprule
\scriptsize Benchmark & \scriptsize DS-V2-Lite-arch Reg. & \scriptsize DS-V2-Lite-arch Far. \\
\midrule
\scriptsize PIQA     & 78.2 & 79.2 \\
\scriptsize ARC-E    & 70.3 & 70.4 \\
\scriptsize ARC-C    & 43.9 & 44.5 \\
\scriptsize HS       & 69.2 & 69.3 \\
\scriptsize WG       & 62.4 & 62.6 \\
\scriptsize MMLU     & 43.3 & 41.7 \\
\scriptsize OpenBook & 41.0 & 40.0 \\
\scriptsize GSM-8K   & 30.9 & 31.0 \\
\scriptsize HEval+   & 26.8 & 23.8 \\
\scriptsize MBPP+    & 48.7 & 49.2 \\
\midrule
\scriptsize Avg      & 51.5 & 51.2 \\
\bottomrule
\end{tabular}
\end{sc}
\end{small}
\end{center}
\vskip -0.1in
\end{table}

\begin{table}[h]
\caption{Computation-communication overlap of all-to-all collectives in overlapped FarSkip-Collective MegatronLM training with EP=8. We evaluate the training of DeepSeek-V2 Lite and a shortened DeepSeek-V3 model with 6 layers.}
\label{tab:overlap_megatron}
\vskip 0.15in
\begin{center}
\begin{small}
\begin{sc}
\setlength{\tabcolsep}{6pt}
\begin{tabular}{lcccc}
\toprule
\scriptsize Method & \multicolumn{3}{c}{\scriptsize all-to-all \% overlap} & ~ \\
\cmidrule(lr){2-4}
& \scriptsize fwd & \scriptsize bwd & \scriptsize Total
& \scriptsize \makecell{end-to-end \\ speedup} \\
\midrule
\scriptsize DS-V2 Lite Regular & \scriptsize 0.0 & \scriptsize 0.0 & \scriptsize 0.0 & \scriptsize 1.0x \\
\scriptsize DS-V2 Lite FarSkip & \scriptsize 87.6 & \scriptsize 89.0 & \scriptsize 88.4 & \scriptsize 1.12x \\
\scriptsize DS-V3 (L=6) Regular & \scriptsize 0.0 & \scriptsize 0.0 & \scriptsize 0.0 & \scriptsize  1.0x \\
\scriptsize DS-V3 (L=6) FarSkip & \scriptsize 92.9 & \scriptsize 84.1 & \scriptsize 88.9 & \scriptsize 1.04x \\
\bottomrule
\end{tabular}
\end{sc}
\end{small}
\end{center}
\vskip -0.1in
\end{table}

In Tab.~\ref{tab:training_comparison}, we study the effect of different distillation techniques and the effect of partial conversion of the model into FarSkip-Collective layers. We start from DeepSeek-V2-Lite MoE and use a short training schedule of 300M tokens using a batch size of $2^{16}$ tokens, 2e-5 learning rate annealed to 1e-5 and test 1)  {\footnotesize\scshape SFT}
 training (Eq.~\ref{eq:sft}) 2) {\footnotesize\scshape KL + Inter. L2} Combining KL with intermediate activation L2 loss (Eq.~\ref{eq:kl} + Eq. \ref{eq:l2}) for which we sweep over different L2 loss coefficients. 3) {\footnotesize\scshape KL} {\tiny\faSnowflake} {\footnotesize\scshape embed.} freezing the embedding and LM-head layers to reduce training instabilities 4) varying batch-sizes but maintaining the same number of training steps. Overall we observe that using the KL objective is the most effective and that freezing the embedding layers does not lead to a significant effect in the model's performance. 
In addition we study the effect of applying FarSkip-Collective to only a subset of the layers, with the layers applied to from the end, i.e., 75\% corresponds to the last 75\% layers of the model converted into FarSkip-Collective layers (cf. Fig.~\ref{fig:sensitivity}). In this settings we still optimize all of the model's parameters and observe that converting fewer layers makes the conversion task significantly easier especially for generation datasets.

We continue to study the effect of the number of modified layers in Fig. \ref{fig:sensitivity} where we use the original checkpoint of Qwen-3-30B MoE and evaluate it under different number of modified layers without training.
We observe modifying the initial layers is more detrimental for performance, which we suspect is the result of two factors. 1) corrupting the early layers will cascade down as corrupted input to later layers and 2) for layer at depth $k$, $f_k$ will have full access to $\frac{k-1}{k}$ of the previous layers via the residual connection, making it less likely to lose critical dependencies for larger $k$. 

\subsection{Explicit Overlapping}
\begin{figure*}[t]
    \centering
    \includegraphics[width=0.4\linewidth]{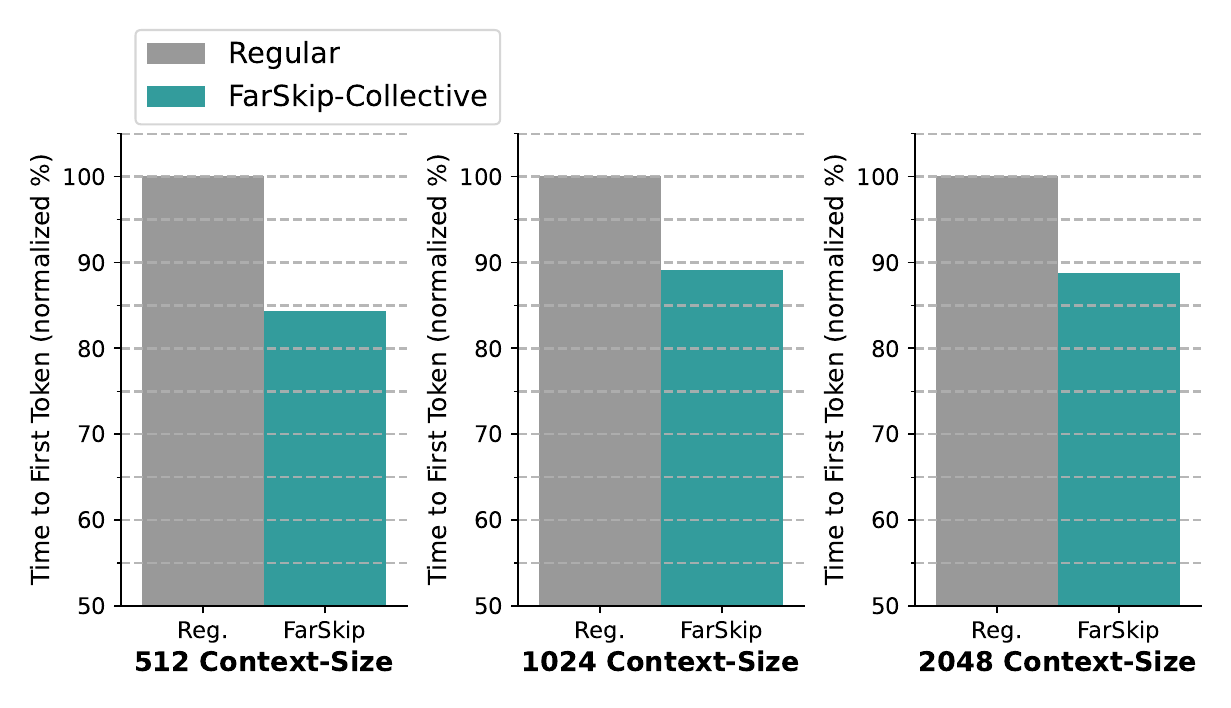}
    \includegraphics[width=0.4\linewidth]{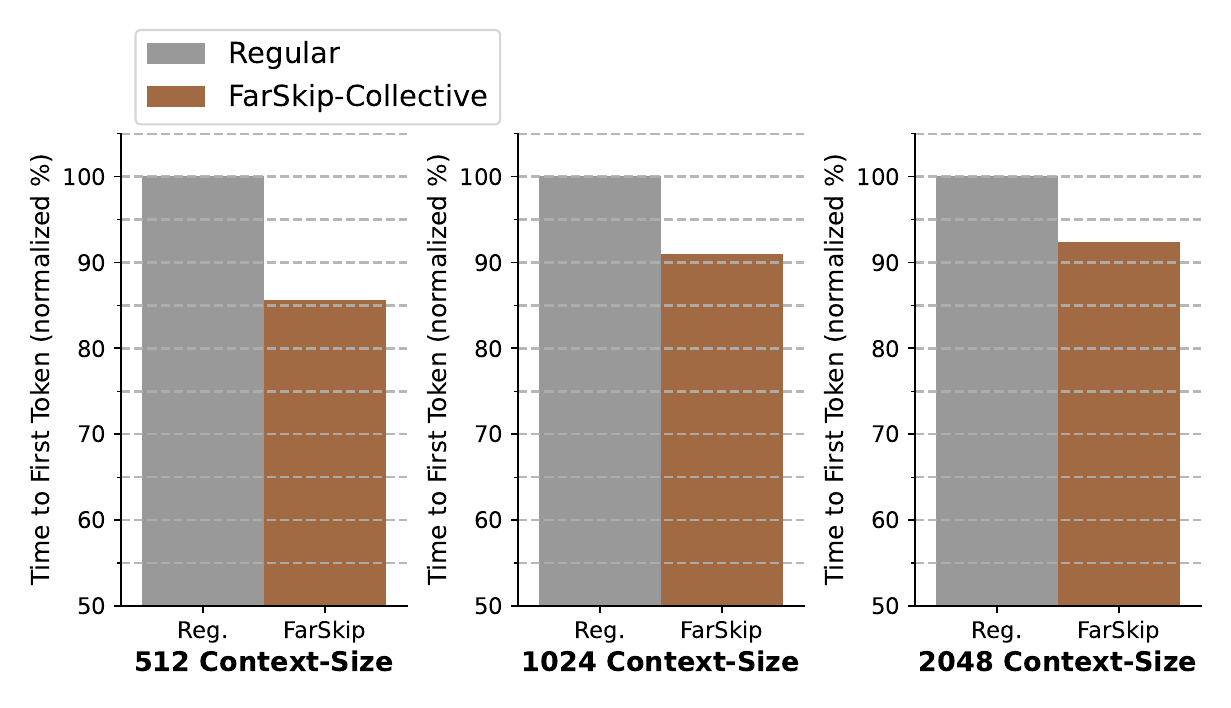}
    
    \vspace{-2mm}
    
    \parbox{0.48\linewidth}{\centering (a) Llama 4 Scout}
    \hfill
    \parbox{0.48\linewidth}{\centering (b) DeepSeek V2}
    
    \caption{Time To First Token (prefill stage) with vLLM inference engine under varying prompt length. Each model is served with EP=8 for the MLP sub-block and TP=8 for attention serving 16 concurrent requests.}
    \label{fig:main}
\end{figure*}

We measure single-node performance of our overlapped training implementation in Megatron-LM in Tab.~\ref{tab:overlap_megatron}, specifically focusing on the all-to-all collectives appearing in the MoE layers. We benchmark training on a node with 8xMI325X and consider two models, DeepSeek-V2 Lite (DS-V2 Lite) (16B) training with a micro-batch size of 4 and global batch-size of 128, and a short DeepSeek-V3 (DS-V3) model with 6 MoE layers (71B) with a micro batch-size of 1. Both models are trained with EP8 and sequence length of 4096. We use the short DS-V3 (L=6) model as it has the same layer dimensions and allows us to study the computation-communication trade-off of the full model while isolating orthogonal factors such as Pipeline-Parallelism (PP). We observe using FarSkip-Collective leads to high degree overlap in both the forward (87.6\%, 92.9\%) and backward pass (89.0\%, 84.1\%) leading to 12\% and 4\% end-to-end speedups in single-node settings for DS-V2 Lite and DS-V3 respectively. This benchmark does not incorporate optimizations such as fused MLA attention that will enable additional acceleration and will make the exposed communication in the model even more critical. \par

\begin{figure}[htbp]
    \centering
    \includegraphics[width=0.95\columnwidth]{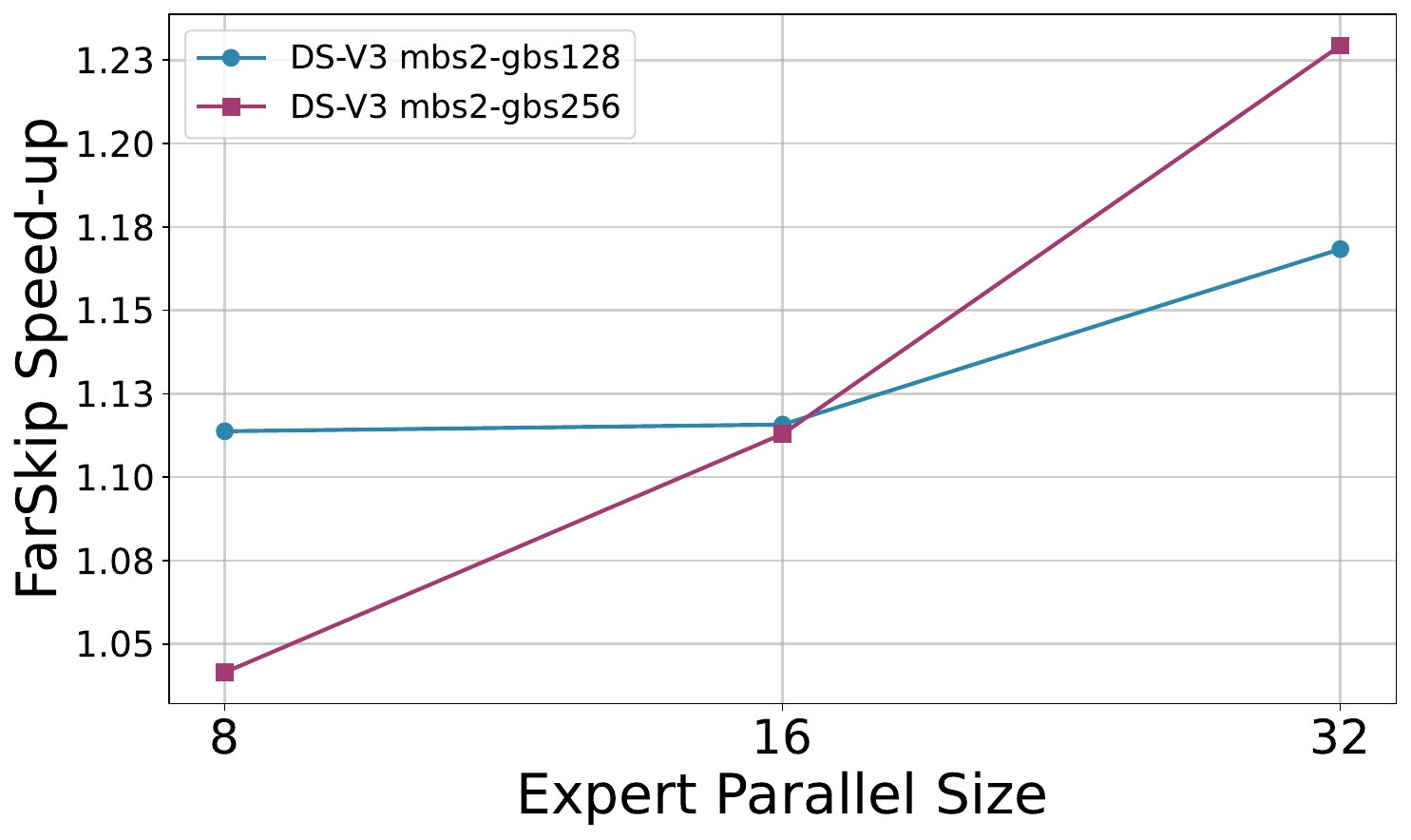}
    \caption{DeepSeek-V3 ($L=6$) FarSkip-Collective training speedup under increasing Expert-Parallelism sizes and different batch size configurations.}\label{fig:strong_scaling}
\end{figure}

We extend the training benchmarking of FarSkip-Collective to multi-node training scenarios on a 4 node system with each node equipped with 8xMI325X GPUs and inter-node communication bandwidth of 400Gbps between GPUs. We study the end-to-end speedup of the DS-V3 $L=6$ model with FarSkip as compared with the regular model training when increasing the number of nodes from 1 to 4 in porportion with the EP size. As we increase the number of nodes and EP we keep the micro (mbs) and global (gbs) batch-sizes fixed (strong-scaling). In Fig.~\ref{fig:strong_scaling} we observe that FarSkip-Collective improvement scales up as we increase the EP size, with EP=32 leading to 1.22x end-to-end training speedup.

For inference with vLLM, we benchmark the prefill phase which has a considerable communication component where we consider the DeepSeek-V2 (235B) and Llama 4 Scout (109B) models. We test both models using a single node with 8xMI300X. In the benchmarking we adopt standard practices and use FP8 quantization and fused-MoE forward kernel (for routed-experts). With this setup we evaluate the Time-To-First Token (TTFT) with different input context lengths (L=512, 1024, 2048), per-device batch size of (BS=2) and (EP=8). For the attention layer the vLLM implementation will mirror the EP size with TP=8. We observe speedups of 8.2\% - 16.8\% and 12.2\% - 18.5\% in both DeepSeek-V2 and Llama-4 in using FarSkip-collective. The smaller number of experts in LLama-4 as compared with DeepSeek-V2 leads to faster computation and makes exposed communication more critical. In addition, we achieve communication overlap of the all-reduce of 95.3\% and 97.6\% for Llama-4 and DeepSeek-V2 (compared with 0\% overlap in regular execution).
In the appendix we share layer execution traces for both training and inference that demonstrate the computation-communication overlap enabled by our implementation.

\begin{figure*}[t]
    \centering
    \includegraphics[width=0.36\linewidth]{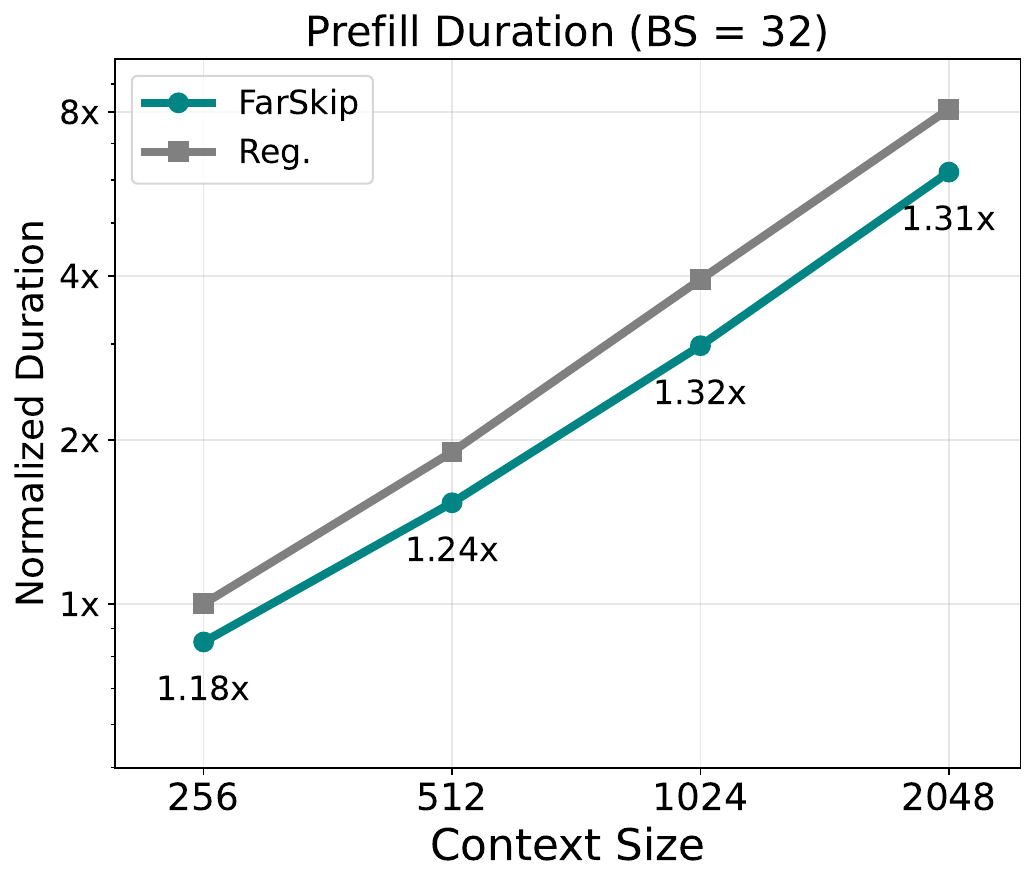}
    \includegraphics[width=0.36\linewidth]{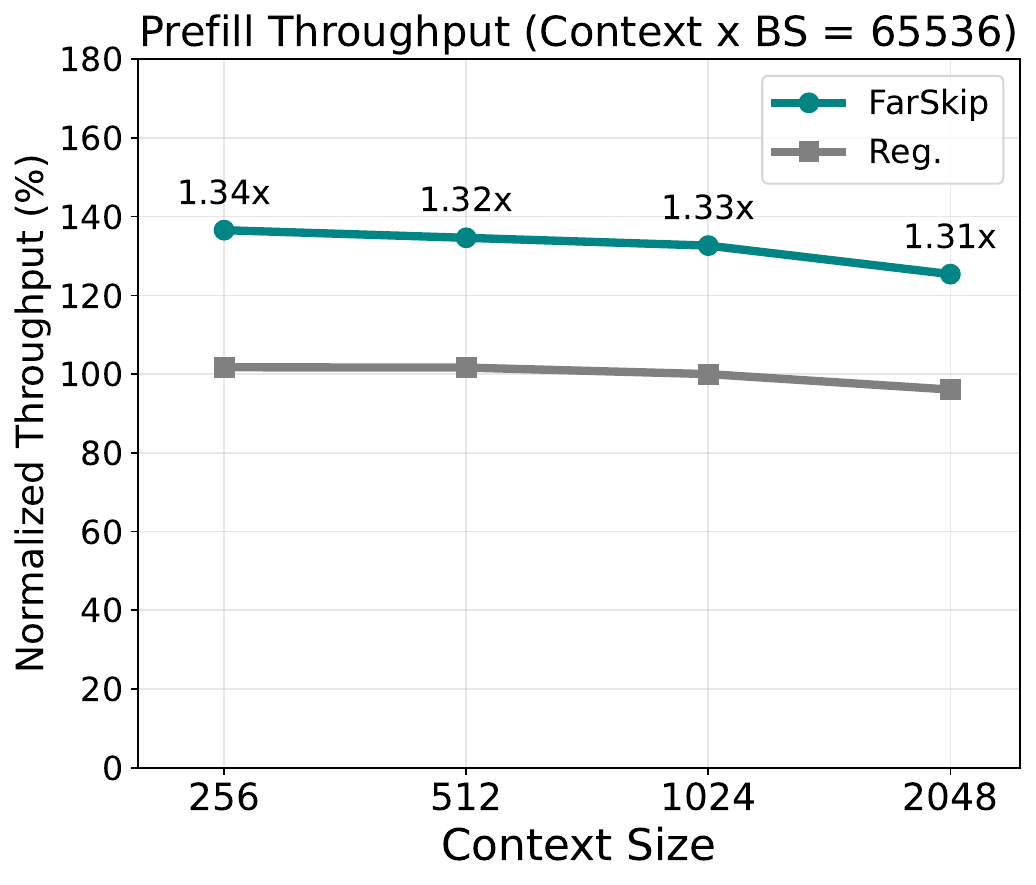}

    \caption{DeepSeek-V3 Time To First Token (prefill stage) with SGLang under varying batch-size and prompt length. Each model is served with EP=8 for the MLP sub-block and TP=8 for attention.}
    \label{fig:ds_v3_prefill}
\end{figure*}

For SGLang inference, we evaluate the optimized DeepSeek-V3 (671B) model architecture equipped with FarSkip-Collective for large-scale MoE inference for both prefill and decoding. In the prefill phase, e.g. benchmarking Time-to-First-Token (TTFT) FarSkip enables up to 1.34x speedup with TP=8, EP=8. 
The prefill stage is compute-bound and Fig.~\ref{fig:ds_v3_prefill} (left) demonstrates linear scaling of the duration of TTFT with the numbers of tokens processed. In both Fig.~\ref{fig:ds_v3_prefill} (left) and (right) we also observe a fairly consistent speedup provided by FarSkip. The duration and the consistent speedup behavior can be attributed to the fact that both the compute-bound computation portion and the bandwidth-bound blocking communication portion scale directly with the number of tokens processed.

\begin{figure}[!h]
\centering
    \includegraphics[width=0.8\linewidth]{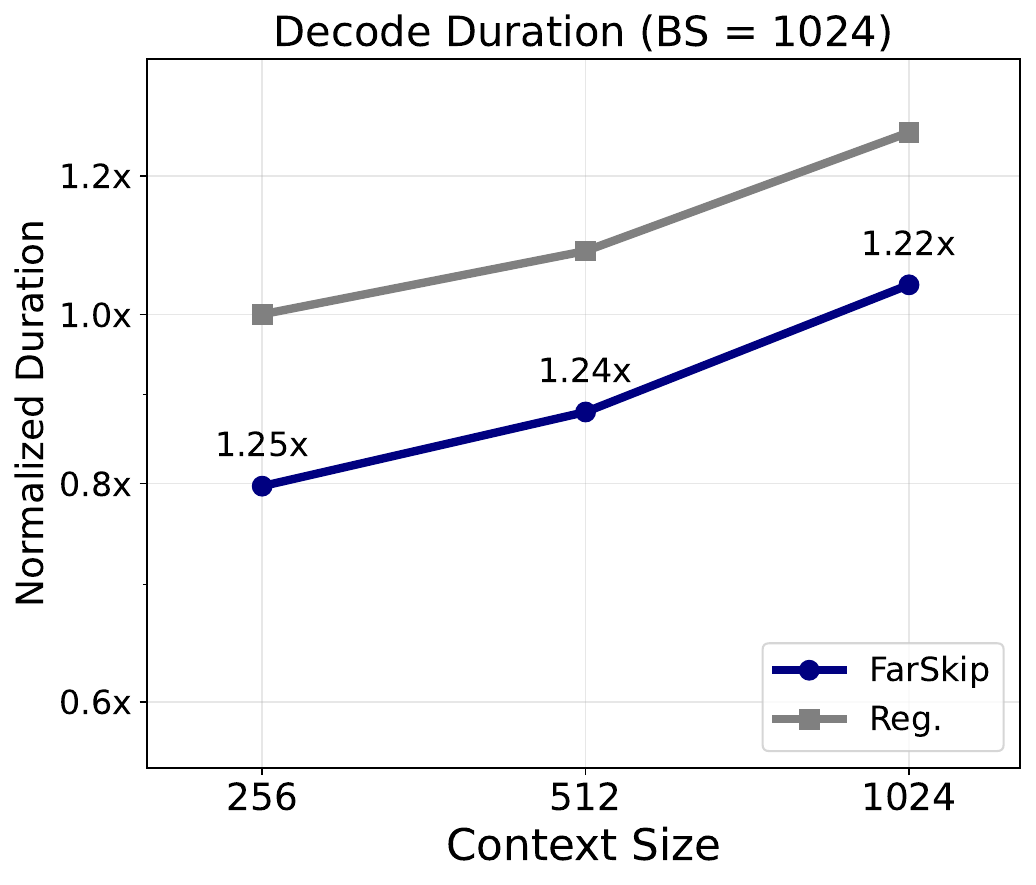}
    \caption{Time Between Tokens (decode stage) with SGLang DeepSeek-V3 for 2-node inference under varying prompt length. Each model is served with EP=16 for the MLP sub-block and TP=16 for attention.}
    \label{fig:ds_v3_decode}
\end{figure}

\begin{figure}[b!]
    \centering
    \includegraphics[width=0.85\columnwidth]{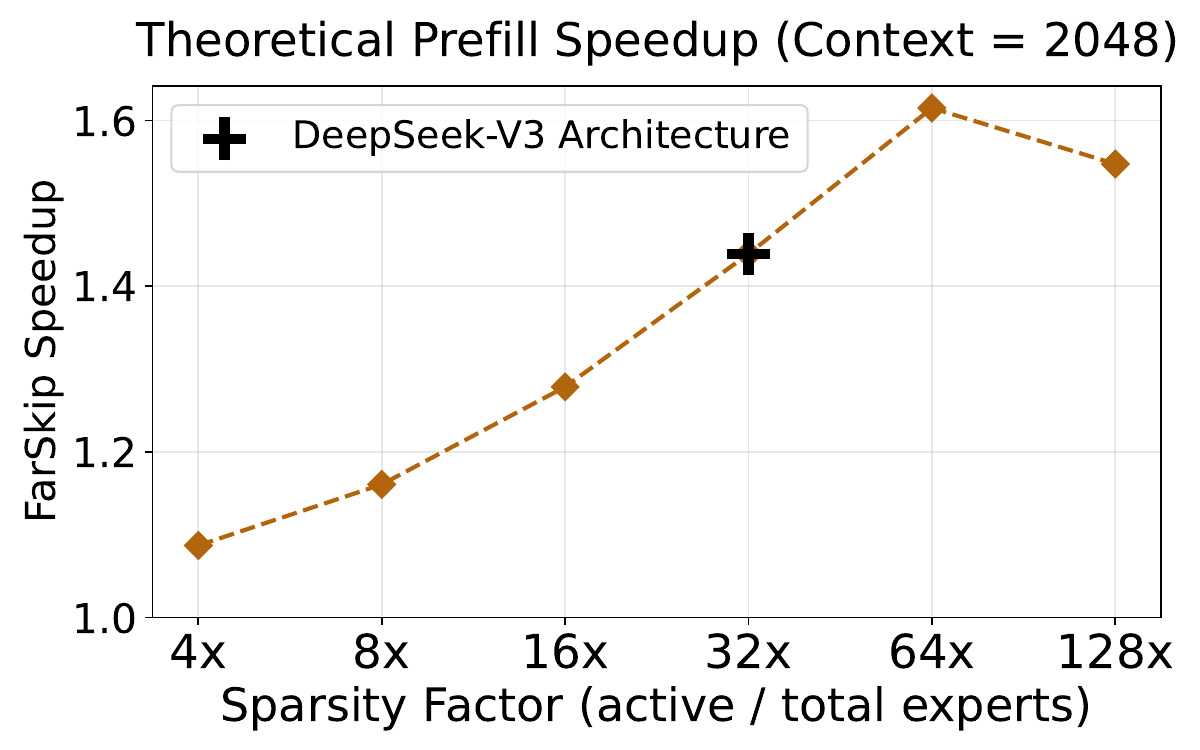}
    \caption{Performance model of FarSkip-Collective prefill speedup under different sparsity levels applied to a Deepseek-V3 architecture.}\label{fig:roofline_main}
\end{figure}

We also create an analytic performance model to experiment with different configurations of models modified with FarSkip-Collective which we present in Fig.~\ref{fig:roofline_main}. Using the theoretical model helps isolating factors such as the presence of optimized kernels and fused operations.
In particular we estimate the speedup provided by FarSkip-Collective under different MoE sparsity levels based on the DeepSeek-V3 architecture (32x sparsity point) and observe that the projected speedup of FarSkip-Collective continues even in sparser settings than DeepSeek-V3 such as 64x and 128x sparsity-level models. We provide additional results and details about the analytic performance model in Appendix \ref{appendix:theoretical_modeling}. 

Unlike the prefill phase, LLM decoding is memory-bandwidth-bound; especially in large MoEs such as DeepSeek-V3. In single-node settings, the large parameter count that needs to be loaded per-GPU, translates to slower decoding and also leads to reduced maximum batch-size due to the limited memory capacity left for the KV cache.
On the communication side, the all-reduce calls are applied to just the newly predicted tokens
which will have smaller message-sizes as compared to prefill, especially with the smaller batches dictated by single-node serving. Together, this makes computation time (that is dictated by memory bandwidth) dominate the single-node large MoE workload as compared with communication.

Nonetheless, in a multi-node set-up FarSkip leads to a significant benefit. Distributing the MoE experts over a larger number of GPUs both directly decreases the computation time and increases the communication time. With wide-EP, the number of experts and parameters per GPU directly decrease and allow for significantly larger batch-sizes. This increases the throughput significantly making large MoE decoding suitable for large-scale distributed setups in general. At the same time, by switching to multi-node serving one relies on scale-out interconnects and the bigger batch-sizes also lead to larger message sizes. Together these changes shift the computation-communication balance making FarSkip-Collective significant in this setting.
To this end, we evaluate DeepSeek-V3 decoding with TP=16, EP=16 in the large-batch setting (BS=1024) on a 2-node 8xMI325X system connected with 8 400Gbps NICs for inter-node communication. In Fig.~\ref{fig:ds_v3_decode} we observe consistent speedup of up to 1.25x with FarSkip under varying prompt lengths as multi-node settings allow for larger batch-sizes.

\section{Related Work}\label{sec:related}
Computation-communication overlap in distributed deep learning traditionally focuses on ``bit-exact'' approaches that retain the mathematical formulation of the model and instead focus on improved execution of the algorithm on hardware. Most existing parallelism techniques aim to achieve minimal exposed communication \cite{zhao2023pytorch,rasley2020deepspeed}. A common theme to achieve overlap is decomposing operators into smaller pieces and scheduling computation and communications in tandem, this includes operator decomposition such as AsyncTP \cite{wang2023overlap,pytorch_async_tp_2024}, and multi-layer pipelines \cite{zhu2025nanoflow,huang2019gpipe,li2023fold3d}.

Closer to our approach are model architectural changes aimed at reducing exposed communication at runtime. 
In the partial formulation front the works of \cite{prabhakar2024kraken} and \cite{lamprecht2025tensor} use un-communicated TP activation shards as input to a sub-block. Nonetheless neither technique can be extended to the MoE settings as the un-communicated shards with EP lead to the full activation in case the selected expert is on the device and a 0 activation otherwise which destabilizing training and inference.
Further both works study the modeling aspect on order of magnitude smaller scales with a restricted set of likelihood tasks compared to our settings. On the implementation side FarSkip-Collective training is the first to enable communication-overlapping in the backward pass and on the inference side coalesces communication to reduce bandwidth by 33\%.
In the outdated formulation the recent work of \cite{zhang2025ladder} uses outdated activations for dense transformers with tensor-parallelism. 
In contrast our work is the first to develop a communication-overlapped architecture for MoEs tackling expert parallelism which entails Dispatch and Combine collectives within the MoE block as opposed to outside the blocks. Our work also demonstrates successful full model conversion (vs. 50\%) and proves out FarSkip-Collective at order-of-magnitude larger scale. 
Other communication-overlapped architecture works include \cite{DLRM19} that designs the model architecture for computation-communication pipelining of the network and computation heavy layers. The works of \cite{black2022gptneox,wang2021gptj} reduce required communication in transformer blocks by designing parallel MLP and attention sub-blocks however neither allows for overlap of the remaining communication. The work of \cite{gunter2024apple} further reduces required model communication for large-scale models via ``track-parallelism'' but also maintains non-overlapped synchronization points.

\section{Conclusion}\label{sec:conclusion}
In this work we present a modified connectivity architecture for MoEs that removes model dependencies between network blocks and therefore has the potential to unhobble blocking communication in their training and execution. We demonstrate the modified architecture has the capabilities to perform well at scale and by developing FCSD we are able to convert a series of state-of-the-art MoE models with increasing scale of up to 109B model parameters efficiently with minor performance degradation. After demonstrating the modified architecture is viable as a replacement of the regular MoE, we move to realize the benefits of FarSkip-Collective by developing optimized training and inference implementations that enable 88.9\% overlapping of all-to-all communication during MoE training and on the inference side enable up to 1.34x speedup in time-to-first-token of large MoEs. Moving forward, by making model execution amenable to overlapping we hope one can rethink existing model architectural configurations and parallelism techniques and explore a larger architecture search space.

\nocite{Bisk2020}
\nocite{allenai:arc}
\nocite{talmor-etal-2019-commonsenseqa}
\nocite{sakaguchi2019winogrande}
\nocite{zellers2019hellaswag}
\nocite{hendryckstest2021}
\nocite{OpenBookQA2018}
\nocite{cobbe2021training}
\nocite{austin2021program}
\nocite{chen2021evaluating}
\nocite{liu2023evalplus}

\bibliography{example_paper}
\bibliographystyle{mlsys2025}

\newpage
\twocolumn
\appendix
\section*{Appendix}

\section{Pre-training results}\label{appendix:pre_training_results}

\begin{figure}[htbp]
    \centering
    \includegraphics[width=0.95\columnwidth]{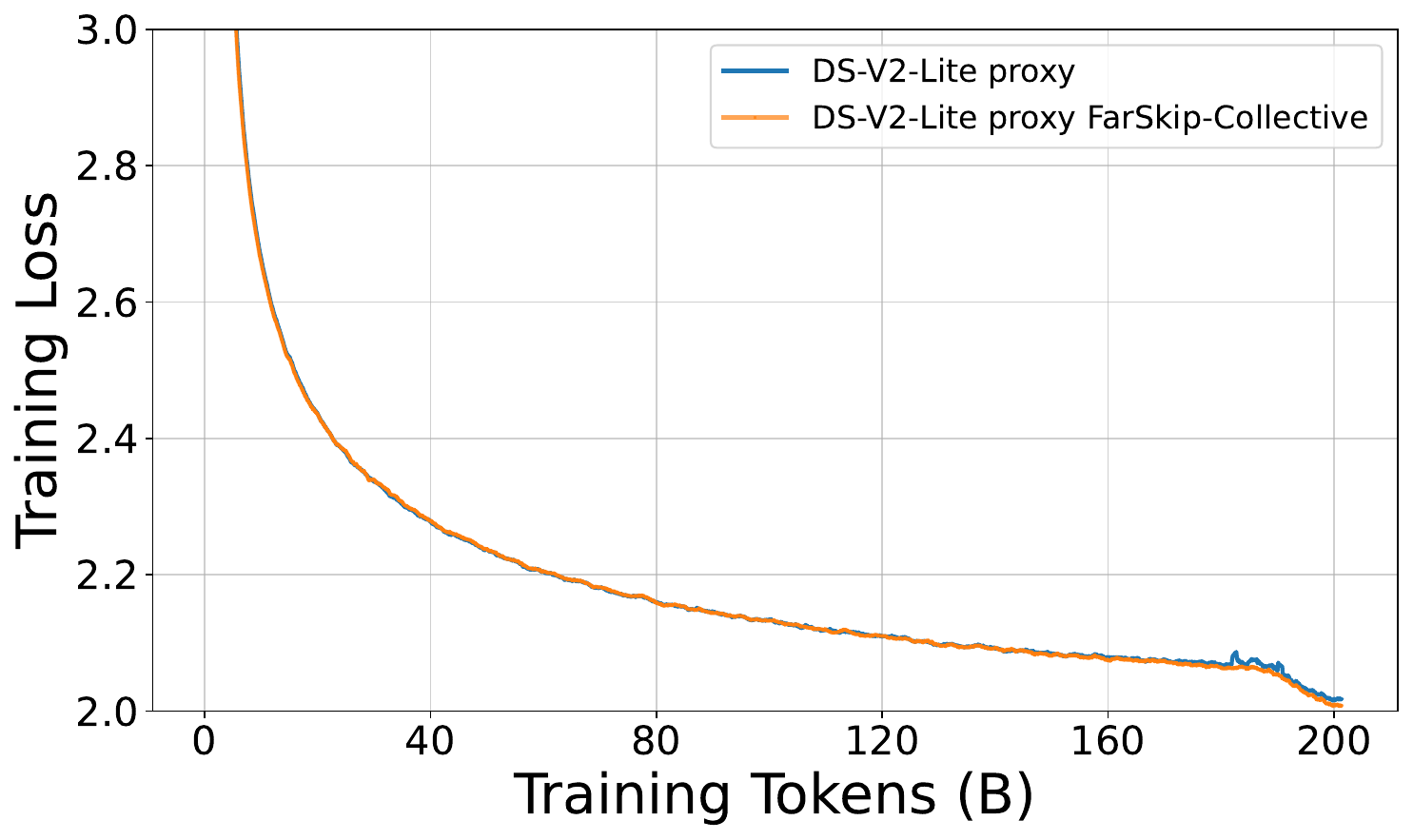}
    \caption{MoE pretraining loss with regular and FarSkip-Collective architectures. We pretrain 16B parameter MoE following the DeepSeek-V2-Lite model architecture for 200B tokens. We observe FarSkip-Collective closely matches the loss of the regular model.}
\end{figure}\label{fig:pretrain_loss}

We test the pre-training of the FarSkip-Collective architecture by pre-training an MoE model from scratch using the DeepSeek-V2 Lite model architecture and configuration (64 routed-experts) in Megatron-LM.  We train the model on 200B training tokens randomly sampled from a larger corpus of high quality mix of publicly available data using a sequence length of 4096 tokens and a batch size of 4,194,304 tokens. With 3B active parameters, this corresponds to 4x simplistic Chinchilla scaling estimate with respect to the model's active parameters \cite{hoffmann2022training}. We repeat the training with the exact settings and seed but with FarSkip-Collective turned off as the regular MoE baseline.  We observe the loss curves of both models matches closely with the FarSkip-Collective model achieving a final training loss of 2.006 as compared with 2.016 averaged over the last 50 steps. For the evaluation we use the same benchmarks as the distillation experiments and observe the two models performance is on par with an average performance gap of FarSkip-Collective of 0.3\%, achieving 51.2\% vs the regular MoE averaged at 51.5\% over the datasets.
This result at 200B token scale with a 16B-A3B model further reinforces the viability of the FarSkip-Collective MoE model modification in addition to the larger-scale self-distillation results in the main paper.

\section{Theoretical Performance Modeling of FarSkip-Collective}\label{appendix:theoretical_modeling}
We develop a theoretical analytical model to analyze the FarSkip-Collective architecture under different scenarios based on hypothetical accelerator performance limits that isolate implementation details such as optimized kernels and fused operations etc.
For this we consider a hypothetical setup of a fake accelerator with the following performance specifications. For computation we assume the dense computations run at 500 TFLOPS for attention and 600 TFLOPS for GEMMs. For communication we assume a GByte/s bus bandwidth for all-reduce of 400 and 300 for all-to-all collectives. We choose relatively smaller compute TFLOPS values for main operations to account for non-dense operations that tend to bring down the cumulative compute utilization estimates. With this set-up we model the prefill stage of inference which remains in the ``compute-bound'' and ``communication-bandwidth-bound'' regimes making the performance analysis with these specifications possible. For these settings we consider an 8-accelerator node setup and use analytical FLOP usage calculations\footnotemark\footnotetext{We estimate FLOPs using the methodology in \href{https://github.com/AI-Hypercomputer/maxtext/blob/fafdeaa14183a8f5ca7b9f7b7542ce1655237574/src/MaxText/maxtext_utils.py\#L454}{github.com/\ \\ \ AI-Hypercomputer/maxtext/src/MaxText/maxtext\_utils.py\#L454} but adapt the calculation for prefill inference (no backward)}
to compute end-to-end run-times of different pieces of the network execution. We compute the estimated runtimes of different components in the model including computation and communication calls based on the proposed limits and the model dimensions specifications. We then compute the regular MoE architecture and FarSkip-Collective architecture by overlapping parts of the communication runtime with the relevant computation components that it can be overlapped with. We summarize the calculation approach for estimated exposed communication runtime for each communication call in the FarSkip-Collective and Regular settings in Tab.~\ref{tab:roofline_comm_schedule}. In our analysis we also assume perfect expert load-balancing per GPU and consider TP=8 EP=8 model parallelisms. We run our analysis for a DeepSeek-V3 architecture for which we vary the levels of sparsity. For the prefill request, we assume BS=32 and context size of 2048. 

\begin{figure}[b]
    \centering
    \includegraphics[width=0.85\columnwidth]{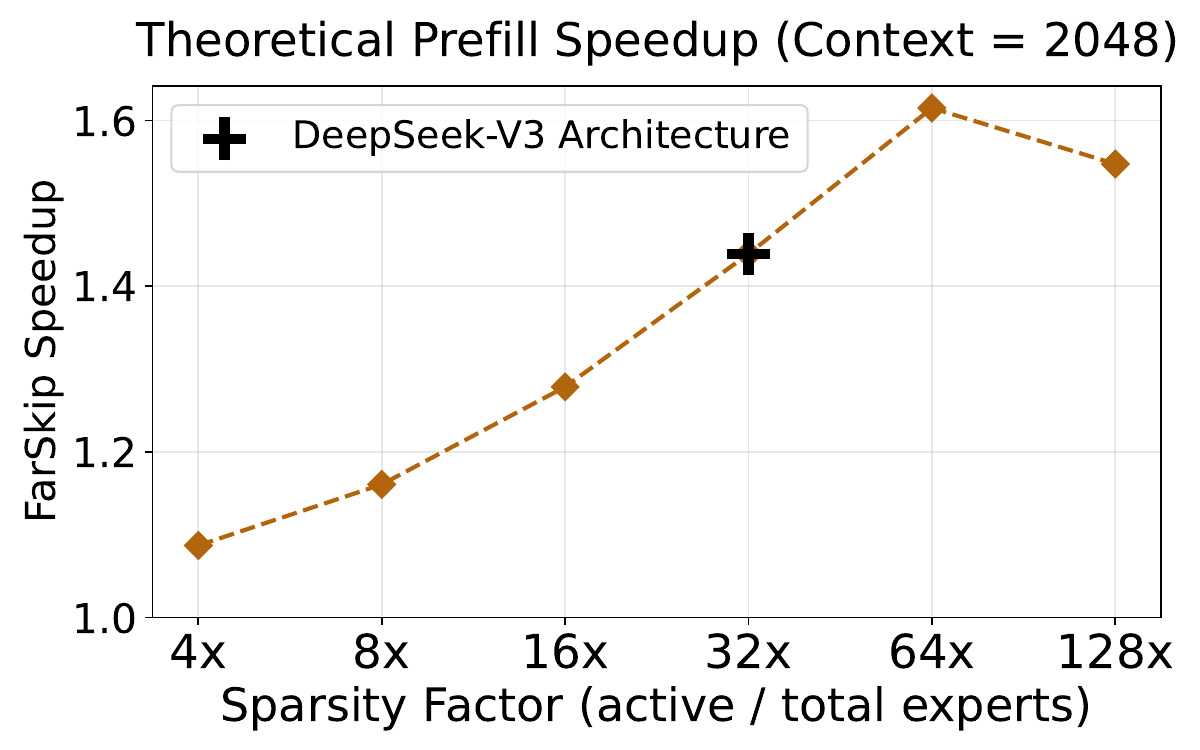}
    \caption*{Figure 9: (reproduced from main) Performance model of FarSkip-Collective prefill speedup under different sparsity levels applied to a Deepseek-V3 architecture.}
\end{figure}

\setlength{\tabcolsep}{3pt}
\begin{table*}[htbp]
\centering
\footnotesize
\caption{%
    Communication-collective operations estimated runtime under FarSkip and Regular MoE settings.
    Under FarSkip, collectives can be overlapped with other computations
    and only the exposed communication latency contributes to the runtime.
}
\label{tab:roofline_comm_schedule}

\begin{tabular}{p{2.4cm} p{1.2cm} p{4.5cm} p{6.5cm}}
\toprule
\thead[c]{Communication\\Collective} & \thead[c]{Settings} & \thead[c]{Contributed Runtime} & \thead[c]{Description} \\
\midrule
{\scriptsize\makecell[c]{Attention \& Shared \\ all-reduce}}
    & {\scriptsize\makecell[c]{FarSkip}}
    & {\fontsize{7.7}{9}\selectfont\makecell[c]{\texttt{min(0, as.comm - routed.runtime)}}}
    & {\scriptsize\makecell[c]{Attention and shared-expert all-reduce is combined into a \\ single call, overlapped over the routed-expert computation, \\ needed as input in the next attention layer.}} \\[14pt]
{\scriptsize\makecell[c]{Routed Expert \\ all-reduce}}
    & {\scriptsize\makecell[c]{FarSkip}}
    & {{\fontsize{7.7}{9}\selectfont\makecell[c]{\texttt{min(0, routed.comm - attn.runtime)}}}}
    & {\scriptsize\makecell[c]{Routed-experts are first used in the shared-expert of the \\ next layer and can be overlapped with attention computation.}} \\[6pt]
{\scriptsize\makecell[c]{Attention \\ all-reduce}}
    & {\scriptsize\makecell[c]{Regular}}
    & {{\fontsize{7.7}{9}\selectfont\makecell[c]{\texttt{attn.comm}}}}
    & {\scriptsize\makecell[c]{Attention all-reduce happens sequentially after the attention layer.}} \\[6pt]
{\scriptsize\makecell[c]{MLP \\ all-reduce}}
    & {\scriptsize\makecell[c]{Regular}}
    & {{\fontsize{7.7}{9}\selectfont\makecell[c]{\texttt{routed.comm}}}}
    & {\scriptsize\makecell[c]{The all-reduce of the routed and shared experts \\ activations happens sequentially after the MLP layer.}} \\
\bottomrule
\end{tabular}

\end{table*}

Below we describe how we model different levels of sparsity for a DeepSeek-V3-like architecture. Recall, DeepSeek-V3 activates 8 + 1 (routed + shared) experts out of 256 + 1 available experts. To modify the sparsity level we consider the scenario of more refined experts but still using 8 + 1 experts activated for each token. For example to model an architecture that is 2x sparser we double the number of experts, while halving the expert dimension, and maintain the number of active experts 8 + 1 (routed + shared). This is realistic as it enables diversity in expert selection while refining the expert size and improving sparsity as seen in recent works such as Kimi-K2 architecture \cite{team2025kimi} which has 1.5x more experts than DeepSeek-V3. With this approach to modeling sparsity we can cleanly maintain the total parameter count (neglecting minimal differences due to the router-gating parameter) while halving the FLOP requirements of token processing of the activated experts. We reproduce Fig.~\ref{fig:roofline_main} below for convenience and observe that FarSkip-Collective provides larger speedups for models that are 2x or even 4x sparser than DeepSeek-V3 which at baseline already independently exhibit larger throughputs due to reduced computations.

We would like to note the limitations of the performance analysis we provide. First we do not properly account for non-dense operations and model all of the dense operations with a simplistic GEMM vs. attention distinction to FLOPS. This does not take into account the potentially-reduced FLOP utilization from smaller GEMMs of more refined experts or non-optimal hardware model dimensions. With regards to FarSkip-Collective speedup, slower GEMMs in the extremely sparse regime will actually lead to continued performance gains as compared to the ``drop-off'' observed at the 128x sparsity level and we should expect a later drop-off as compared with the simple performance model. Lastly we do not account for potential reduction in compute efficiency that can result from communication overlapping due to competing hardware resources.

\begin{figure}
    \centering
    \includegraphics[width=0.85\linewidth]{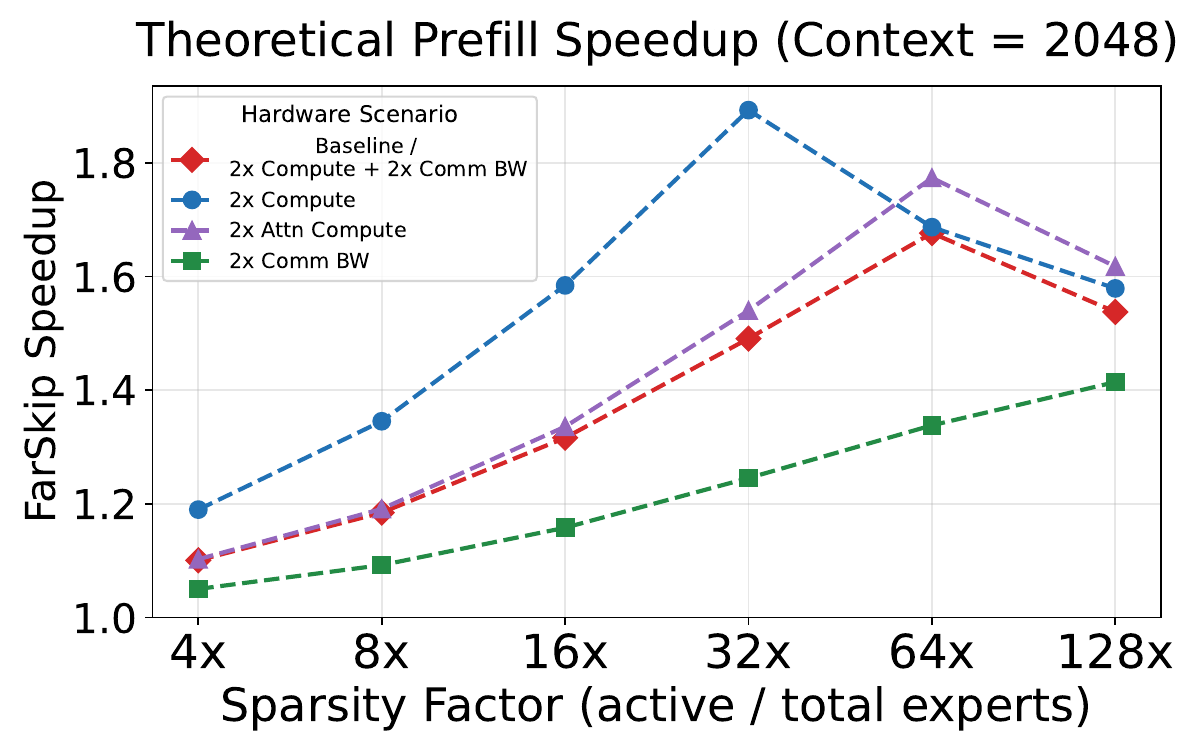}
    \caption{Theoretical speedup of FarSkip-Collective prefill under different compute and communication scenarios for a range of sparsity levels modified from the DeepSeek-V3 architecture (original model @ 32x sparsity factor).}
    \label{fig:roofline_scenarios}
\end{figure}

In Fig.~\ref{fig:roofline_scenarios} We consider the effects of different hypothetical hardware/algorithmic settings such as increased or reduced computation or communication speeds.
In particular we evaluate the scenarios of 1) 2x faster computation, 2) 2x faster attention computation (e.g. suppose in the very sparse settings the small shapes of the MoE multiplication lead to lower utilization compared with attention) 3) 2x faster communication and 4) 2x faster communication and 2x computation. For the compute-bound, communication-bandwidth-bound model the ``2x compute + 2x communication'' scenario coincides with the baseline scenario as everything scales down proportionally while absolute speeds double in both the FarSkip-Collective and baseline models.

\section{Optimized Training: Overlapped Backward Call Implementation}\label{appendix:backward_pass}
In this section we provide more details on our communication-computation overlapped implementation of FarSkip training, specifically how we achieve computation-communication overlapping in the backward pass using two techniques. Recall, in the MoE layer with EP all-to-all communication collectives appear in the Dispatch and Combine operations within the MLP block. The gradients of the Dispatch and Combine operations are themselves all-to-all collectives which we would like to overlap. The issue is that with autograd, one does not control the node traversal order in the backward pass, risking a race condition or no overlap at all. Instead we propose two innovative techniques that put together, enable us to achieve overlap cleanly. 1) we implement an async-safe all-to-all custom autograd function that triggers a synchronization before future processes access its output tensors which avoids race conditions. and 2) we use the Sequence Numbers in PyTorch's internals API to reprioritize node traversal of autograd processing in order to delay the synchronization triggers for as long as possible to ensure sufficient time for communication to run and overlap. 
\par To implement the async-safe all-to-all custom autograd function we create an autograd class for async-all-to-all with a stateful dictionary that stores both the forward and backward communication handles produced by async communication in PyTorch. During the forward pass, the forward all-to-all handles are being generated by the collective in async mode; while the backward-all-to-all communication handles do not exist yet but they have a dedicated keys in the layer's stateful dictionary. When \texttt{backward()} is called on the operator, it runs backward via another all-to-all collective call in async mode and returns a communication handle which we use to populate the state dictionary. This enables us to store and later access the backward handle while not explicitly calling the backward function of the async-all-to-all that gets called implicitly by autograd. We then implement a backward-hook that takes as input the async-safe autograd function object along with its stateful dictionary and will trigger synchronization of the backward handle when the hook fires. 
We then attach the hooks to the input tensors of the all-to-all operators via PyTorch's  \texttt{register\_full\_backward\_hook}. At the time we register the hook, the state dictionary is empty but before the hook will be fired async-all-to-all backward will take place and the backward handle keys will be populated.  The hook + stateful dictionary approach ensures that we can implicitly pass communication handles to the hook and trigger a synchronization of the handles before the processing of any gradients that will use the tensors produced by the async-all-to-all in the backward call. 
Overall this makes it possible to run the all-to-all communication in the backward pass asynchronously while ensuring the relevant gradients are ready when accessed.\par 

Nonetheless async-safe all-to-all gradient is not sufficient to achieve significant overlapping of the backward pass. As we are optimizing overlap with explicit ordering in the forward pass, the communication in the forward pass is implemented such that as soon as computation produces all the outputs that are consumed by a communication collective, the communication call will launch. This maximizes the overlapping window in the forward pass by starting the communication as early as possible. In the backward pass, however, this leads to the opposite effect as the inputs to communication calls will now be launched immediately after the backward communication call and the handles are forced to be synchronized immediately after launching which causes a communication bubble by synchronizing too early in the backward pass. \setcounter{footnote}{0} To resolve this, we ``hijack'' the priority ordering of \texttt{torch.autograd} via the Sequence Number PyTorch autograd's internal implementation\footnotemark \footnotetext{See ``forward-backward correlation'' discussion of Sequence Number autograd internals in paragraph below the anchor {\scriptsize \href{https://docs.pytorch.org/docs/stable/autograd.html\#torch.autograd.profiler.emit_nvtx}{https://docs.pytorch.org/docs/stable/autograd.html\#torch.autograd.profiler.emit\_nvtx}}}. In torch autograd, the computational graph will be processed according to a topological sorting algorithm of the dependencies between nodes. However, when multiple nodes are ready for processing at the same time, autograd uses Sequence Numbers to select the first node to process which are typically ordered based on node's chronological creation during forward. As we explained, this leads to the undesired effect where async communication is synced as soon as it launches with our optimized implementation of the forward pass. However with the connectivity of FarSkip-Collective, the dependency drops mean that one can actually process an entire sub-block's backward pass autograd nodes before reaching a dependency barrier on the input to the communication call. Harnessing this, we manually reprioritize the autograd priority queue to prioritize nodes in the sub-block's computation and de-prioritize the processing of the computations leading to the input of the communication call by reassigning them custom Sequence Numbers. With this reassignment those nodes will be launched for backward computation only after the non-dependent sub-block computations took place, allowing for large overlap opportunities before the synchronization points get triggered. Overall with this approach we can flexibly implement different overlapped backward settings and control the execution cleanly without handwriting massive backward autograd functions that will require manually computing every gradient of every sub-block of a layer (attention, MLP and MoE weights and activations).

\section{FarSkip-Collective Layer Traces}
We present excerpt layer traces of explicitly overlapped FarSkip Models during training and inference.

\begin{figure*}[htbp]
    \centering
    \includegraphics[width=1.9\columnwidth]{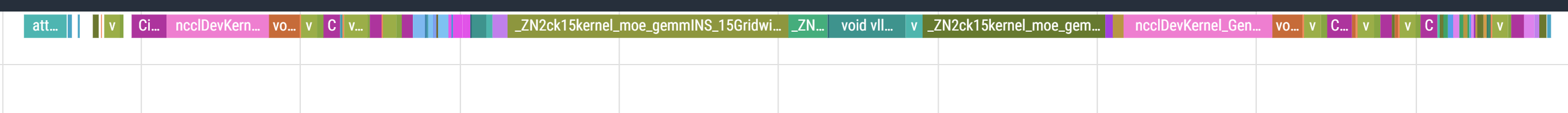}
    
    \vspace{0.3cm} 
    
    \includegraphics[width=1.9\columnwidth]{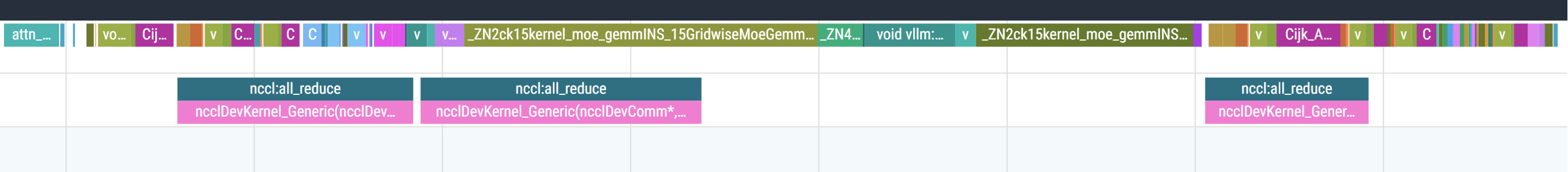}
    \caption{\textbf{DeepSeek-V2 vLLM prefill inference layer execution} (Top) regular connectivity (Bottom) FarSkip-Collective. In the bottom figure the all-reduce collectives are overlapped during the attention and MoE sub-blocks by running asynchronously on a 2nd Hardware Queue.}
    \label{fig:stacked}
\end{figure*}

\begin{figure*}[htbp]
    \centering
    \includegraphics[width=1.9\columnwidth]{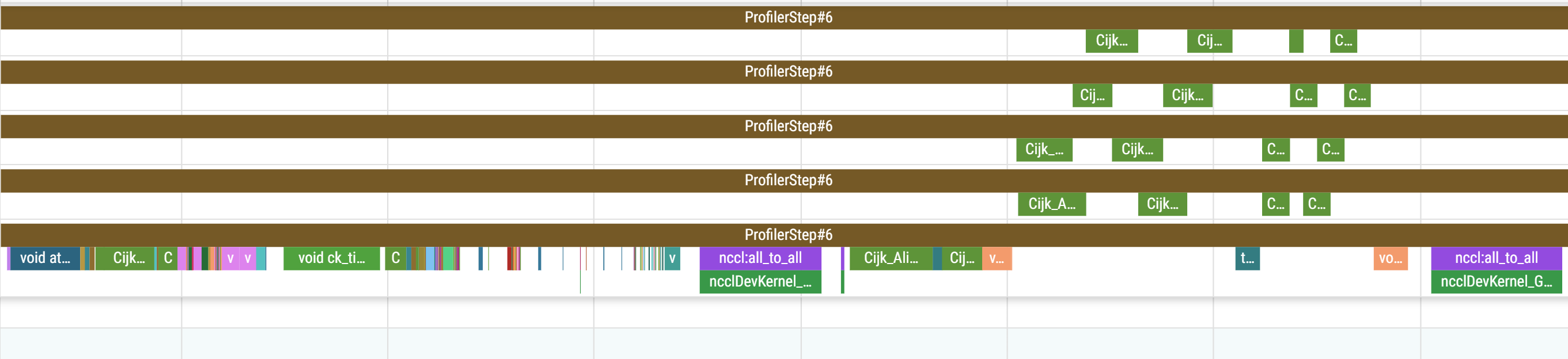}
    
    \vspace{0.3cm}
    
    \includegraphics[width=1.9\columnwidth]{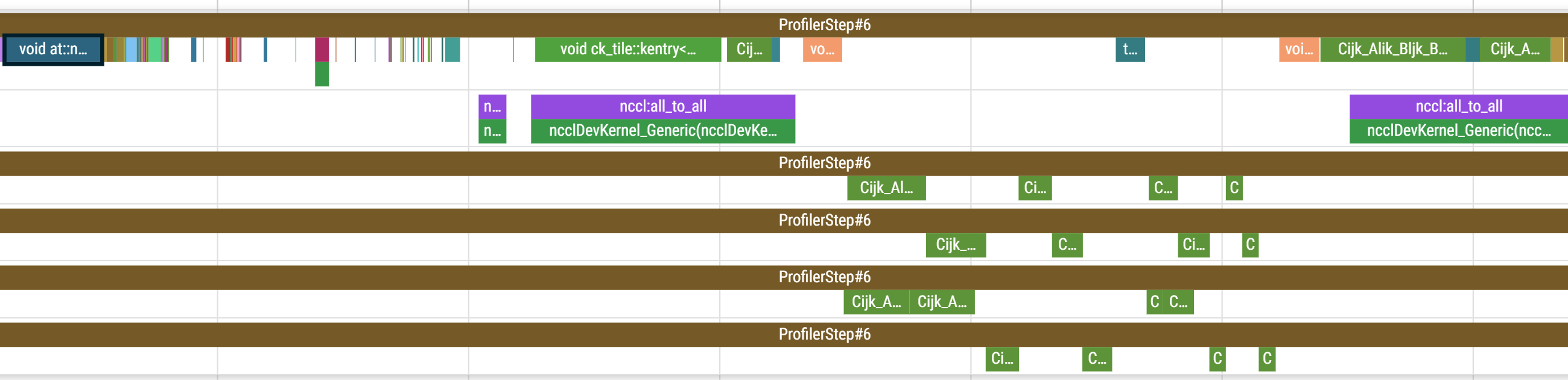}
    \caption{\textbf{DeepSeek-V2-Lite pre-training \emph{forward}-pass layer execution} (Top) regular connectivity (Bottom) FarSkip-Collective. In the bottom pane,  all-to-all communication is overlapped with computation, the first call corresponds to Dispatch which gets overlapped with the core-attention computation. In the second call, the all-to-all corresponds to Combine and is overlapped with the shared-experts and the next layer's $q,k,v$ computation for attention.}
    \label{fig:stacked}
\end{figure*}

\begin{figure*}[htbp]
    \centering
    \includegraphics[width=1.9\columnwidth]{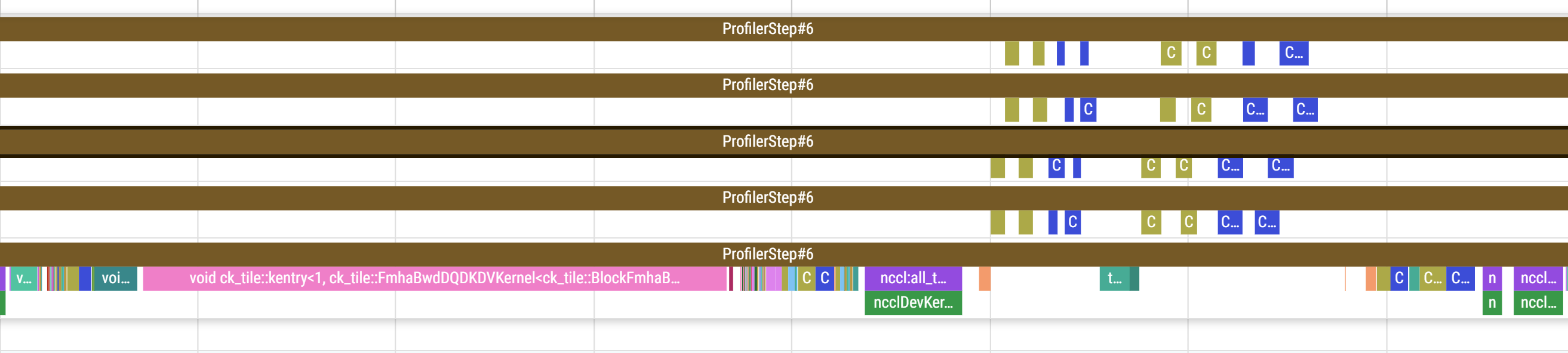}
    
    \vspace{0.3cm}
    
    \includegraphics[width=1.9\columnwidth]{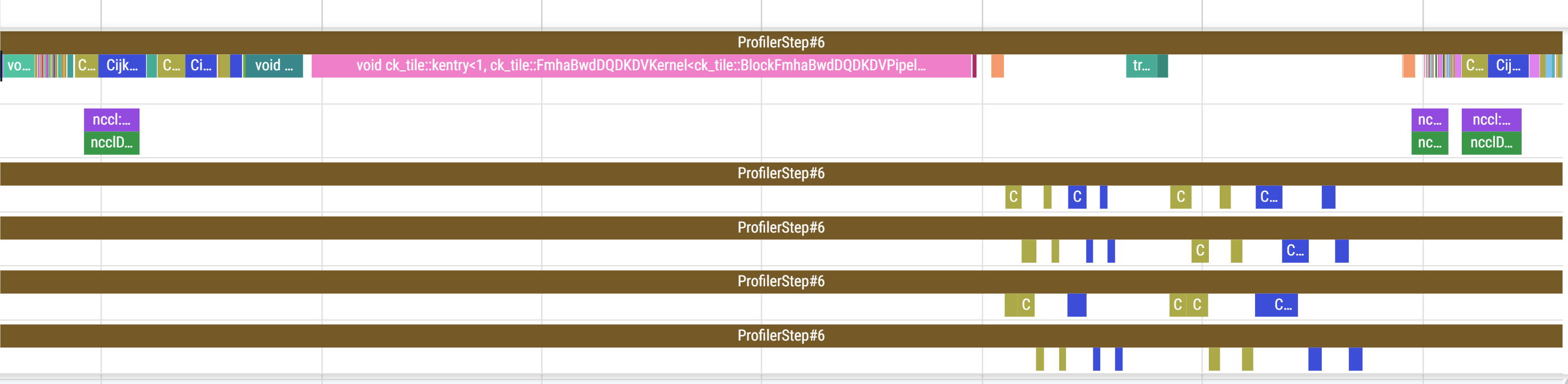}
    \caption{\textbf{DeepSeek-V2-Lite pre-training \emph{backward}-pass layer execution} (Top) regular connectivity (Bottom) FarSkip-Collective. The backward-pass operator execution order is ``hijacked'' from the default \texttt{torch.autograd} Sequence Number ordering to re-order operations for overlap. In particular, routed-expert backward computation launches immediately after the finished synchronization point of the Combine all-to-all backwards gradient and the Dispatch gradient launches before the gradient of the first part of attention ($q,k,v$ calculation) to allow for overlapping with it before synchronization.}
    \label{fig:stacked}
\end{figure*}

\end{document}